%% file: main.tex
\documentclass[lettersize,journal]{IEEEtran}

\usepackage{amsmath}

\usepackage{amssymb}
\usepackage{amsfonts}
\usepackage{graphicx}
\usepackage{subcaption}
\usepackage{booktabs}
\usepackage{hyperref}
\usepackage{colortbl}
\usepackage{multirow}
\usepackage{placeins}
\usepackage{algorithm}
\usepackage{enumitem}
\usepackage{algpseudocode}
\usepackage{array}
\usepackage{url}
\usepackage{cite}

\providecommand{\Description}[1]{}

\begin{document}

\title{Drift-Based Policy Optimization: Native One-Step Policy Learning for Online Robot Control}

\author{
Yuxuan Gao$^{1,*}$, 
Yedong Shen$^{1,*}$, 
Shiqi Zhang$^{1}$, 
Wenhao Yu$^{1}$, 
Yifan Duan$^{2}$, 
Jia Pan$^{3}$, 
Jiajia Wu$^{3}$, 
Jiajun Deng$^{1,\dagger}$, 
Yanyong Zhang$^{1,\dagger}$~\IEEEmembership{Fellow,~IEEE}
\thanks{$^{*}$These authors contributed equally.}
\thanks{$^{\dagger}$Corresponding authors.}
\thanks{$^{1}$University of Science and Technology of China, Hefei 230026, China. 
Emails: {\{yuxuangao, sydong2002, zhangshiqi\_1127, wenhaoyu\}@mail.ustc.edu.cn, \{dengjj, yanyongz\}@ustc.edu.cn}}
\thanks{$^{2}$LYNSENSE, Shanghai, China. Email: duanyifan@lynsense.net}
\thanks{$^{3}$iFLYTEK, Hefei 230088, China. 
Emails: \{jiapan, jjwu\}@iflytek.com}
}

\maketitle

\begin{abstract}
\input{secs/0_abstract}
\end{abstract}


\begin{figure*}[t]
  \centering
  \includegraphics[width=\textwidth]{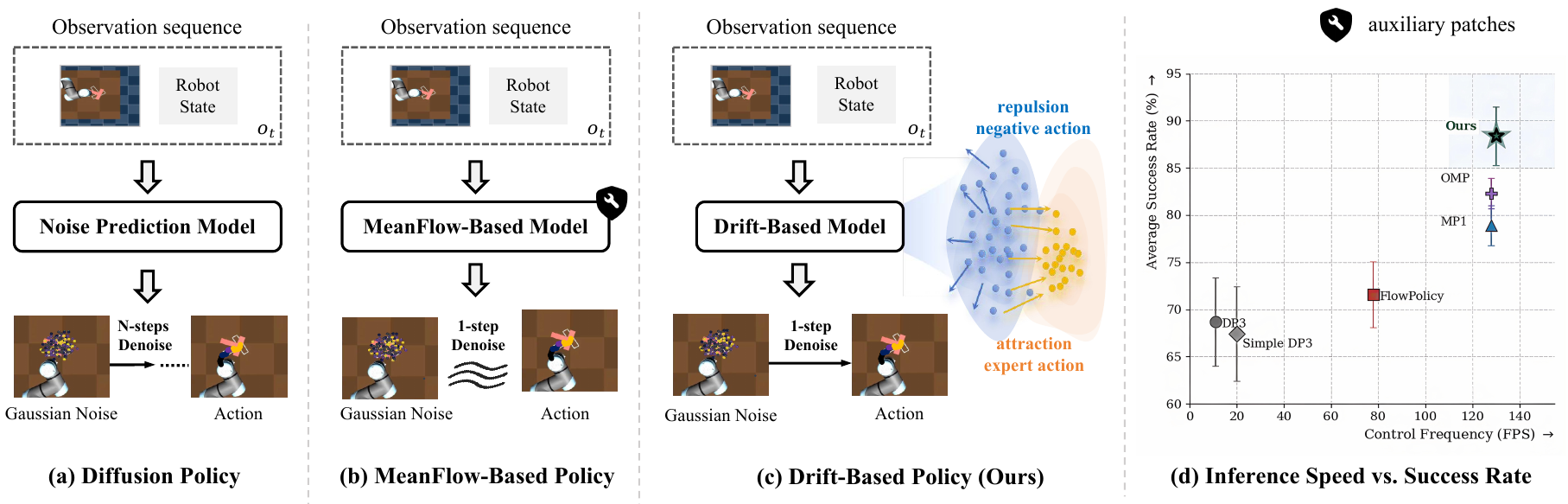}
  \caption{Generative policy paradigms for robot control. (a) Multi-step diffusion policies rely on iterative denoising at inference. (b) One-step mean-flow policies generate actions in one pass with auxiliary corrections. (c) Drift-Based Policy internalizes attraction-repulsion refinement during training, yielding a native one-step generator. (d) Our method achieves the best average success rate and control frequency on Adroit and MetaWorld against established generative baselines.}
  \Description{A four-panel figure: (a) Multi-step Diffusion Policy. (b) One-Step MeanFlow-Based Policy. (c) Drift-Based Policy showing training refinement via attraction/repulsion. (d) A scatter plot showing Drift-Based Policy achieving the highest success rate and inference frequency (FPS) compared to DP3, MP1, and OMP.}
  \label{fig:intro_comparison}
\end{figure*}

\input{secs/1_intro}
\input{secs/2_related}
\input{secs/3_preliminaries}
\input{secs/4_method}
\input{secs/5_experiments}

\input{secs/6_conclusion}
\clearpage
\bibliographystyle{IEEEtran}
\bibliography{ref}

\newpage
\onecolumn
\appendices
\input{secs/supplementary_content}

\end{document}

%% file: secs/0_abstract.tex
Diffusion policies effectively model multimodal action distributions for robotic manipulation, but their iterative denoising requires tens to hundreds of network function evaluations (NFEs) for each control prediction, limiting their applicability to high-frequency closed-loop control and online reinforcement learning (RL). We present a two-stage framework for native one-step generative policies that transfers iterative refinement from inference to training. First, Drift-Based Policy (DBP) uses a fixed-point drifting objective to internalize corrective dynamics into the model parameters, producing multimodal action chunks with a single network evaluation by design. Second, Drift-Based Policy Optimization (DBPO) augments the pretrained backbone with a stochastic interface that provides exact conditional rollout likelihoods for PPO-style on-policy updates while preserving 1-NFE deployment. On the 12-task Diffusion Policy suite, DBP improves the average success rate from 0.79 to 0.83 while reducing inference from 100 NFEs to 1. Across 37 point-cloud manipulation tasks, DBP achieves an average success rate of 88.4\%, surpassing the leading 1-NFE baseline OMP at 82.3\%. DBPO further improves pretrained one-step policies through stable online fine-tuning on RoboMimic and D4RL. On a physical dual-arm UR5 platform, DBP achieves 123/150 successes (82\%) with an average end-to-end latency of 9.5~ms, compared with MP1's 89/150 successes (59\%) under the same setup. Code has been released on \url{https://github.com/YuxuanGao0822/DBPO}.

%% file: secs/1_intro.tex
\section{Introduction}
\label{sec:intro}

Robotic manipulation requires policies capable of executing complex visuomotor tasks across diverse conditions. A primary challenge arises from action multimodality: for a given observation, multiple valid action sequences can exist due to task ambiguity, demonstration diversity, or environmental stochasticity. This challenge has motivated recent research toward expressive generative policies that model actions as conditional distributions rather than deterministic mappings. Multi-step generative policies, exemplified by the Diffusion Policy~\cite{Chi-RSS-23} and DP3~\cite{Ze2024DP3}, demonstrate strong performance on challenging manipulation tasks through iterative refinement during inference. However, this iterative mechanism necessitates multiple network evaluations per action---often tens to hundreds of denoising or transport steps---which incurs substantial inference latency. This latency proves prohibitive for high-frequency closed-loop control and severely restricts online reinforcement learning~\cite{ren2025dppo,zhang2025reinflowfinetuningflowmatching,zou2026stepenoughdispersivemeanflow}.

To address this bottleneck, recent studies explore one-step or few-step generative policies, with the goal of reducing the process of robotic control to a single network function evaluation. These approaches typically fall into two categories. Methods based on diffusion acceleration~\cite{prasad2024consistency,wang2024onestepdiffusionpolicyfast,consistencypolicy2024,twostep2025,onedp2025} obtain one-step behavior by distilling, compressing, or accelerating pretrained multi-step generators in a consistency style, thereby inheriting the capabilities of the teacher model while retaining a dependency on multi-step pretraining. Mean-flow-based policies~\cite{sheng2025mp1,zou2025dm1,zhan2026mvp,geng2025meanflow} achieve one-step deployment without distillation, yet they rely on auxiliary quality-preserving or corrective mechanisms---such as dispersive losses, directional alignment, or instantaneous-velocity constraints---to maintain performance under strict 1-NFE constraints.

% Although both categories demonstrate the practical utility of efficient inference, each entails a distinct trade-off. Distillation-based routes leverage robust teacher models but remain tied to multi-step pretraining, whereas correction-based routes stabilize strict 1-NFE generation through additional objective terms. In this work, the term native one-step refers to formulations in which 1-NFE inference arises directly from the training objective itself, rather than from post-hoc acceleration or objective coupling. This distinction reveals a remaining gap in current policy learning: a formulation that jointly delivers high policy quality and one-step efficiency as an intrinsic property of the backbone.
Although both categories demonstrate the practical utility of efficient inference, each entails a distinct trade-off. Distillation-based routes leverage robust teacher models but remain tied to multi-step pretraining, whereas correction-based routes stabilize strict 1-NFE generation through additional objective terms, often introducing extra optimization constraints. In this work, the term native one-step refers to formulations in which 1-NFE inference arises directly from the training objective itself, rather than from post-hoc acceleration or objective coupling. This distinction reveals a remaining gap in current policy learning: a formulation that jointly delivers high policy quality and one-step efficiency as an intrinsic property of the backbone. Addressing this gap requires rethinking policy learning from a training-centric perspective, where efficiency and performance are co-designed rather than sequentially approximated.

We address this gap by building upon Drifting Models~\cite{deng2026generative}, a generative principle designed to be one-step by construction. Unlike diffusion and flow methods that perform iterative refinement during inference, drifting shifts the refinement process entirely to the training phase. During the learning phase, the generator progressively internalizes corrective behaviors through the training-time evolution of the pushforward distribution, such that post-training inference requires only a single forward pass to produce high-quality samples. Mechanistically, this design differs from post-hoc acceleration: low-latency deployment is achieved through training dynamics rather than through compression or distillation. For robotic control, this distinction proves practically significant because latency is guaranteed by the formulation of the policy itself, yielding a predictable 1-NFE deployment path without reliance on additional corrective modules.

This principle is instantiated as the Drift-Based Policy (DBP), a native 1-NFE policy backbone for observation-conditioned action generation in robotic manipulation. The DBP is designed around three practical requirements for deployment-oriented policy learning. First, it utilizes a drifting formulation specialized for sequential action-chunk prediction under observation conditioning. Second, it provides native support for heterogeneous sensory modalities, including low-dimensional states, RGB images, 3D point clouds, and multi-camera inputs. Third, it strictly preserves 1-NFE inference under closed-loop control. This design establishes drifting as a practical native one-step policy backbone that achieves high performance through native generative principles rather than through post-hoc compression of multi-step models.

However, a robust one-step backbone alone remains insufficient for a complete control framework. As an offline imitation learner, the DBP primarily reproduces the demonstration distribution and can be limited when the improvement of returns requires behavior beyond demonstration support. Online RL provides complementary benefits by directly optimizing task rewards, improving recovery from off-demonstration states, and expanding effective state-space coverage in evaluation settings. To bridge this gap, we introduce Drift-Based Policy Optimization (DBPO), which extends the DBP to online reinforcement learning while preserving the 1-NFE deployment efficiency of the backbone. The primary challenge involves concurrently preserving the multimodal expressiveness of the pretrained one-step generator, supporting exact action-likelihood evaluation for standard on-policy optimization algorithms, and maintaining strict 1-NFE inference during deployment. This challenge is addressed through a minimal stochastic interface that enables exact likelihood computation for on-policy updates while keeping the deterministic generation path unchanged during deployment. Consequently, DBPO performs on-policy RL updates without sacrificing the native one-step property established during offline training.

The proposed framework is evaluated along two complementary axes. First, the DBP is compared against the multi-step Diffusion Policy to verify that native one-step generation can match or exceed multi-step performance at a fraction of the inference cost. Across the simulation suite of the Diffusion Policy (12 tasks), the DBP improves the family-level average score from 0.79 to 0.83 while reducing the inference cost from 100 NFEs to 1. Second, the proposed framework is benchmarked against competitive 1-NFE baselines across both offline imitation learning and online fine-tuning settings. For point-cloud manipulation across Adroit and Meta-World, the DBP establishes a state-of-the-art success rate of 88.4\%, consistently outperforming representative 1-NFE baselines. The comprehensive evaluation further includes online RL benchmarks on RoboMimic and D4RL, where DBPO improves upon a strong one-step offline initialization via PPO fine-tuning while strictly preserving 1-NFE deployment. Additionally, real-world deployment on a physical dual-arm UR5 setup yields a 75\% success rate at $105.2\,\mathrm{Hz}$, confirming its practical feasibility for high-frequency control.

The contributions of this work are summarized as follows:
\begin{itemize}[itemsep=0pt, parsep=0pt, topsep=2pt]
    \item We introduce the Drift-Based Policy (DBP), a native one-step generative policy for robotic control. This approach shifts iterative refinement from inference to training via fixed-point drifting objectives, achieving a 1-NFE deployment by design while preserving the capacity for multimodal action modeling.
    \item We propose Drift-Based Policy Optimization (DBPO), an online reinforcement learning framework built upon DBP. This framework overcomes the performance ceiling and spatial generalization limits of offline imitation learning by enabling exact-likelihood on-policy updates while preserving a deterministic one-step deployment path.
    \item We present comprehensive empirical validation across simulation, real-world deployment, and both offline and online learning regimes. The results demonstrate that DBP achieves competitive or superior performance compared to multi-step diffusion-based policies while reducing inference to a single forward pass, and consistently outperforms existing 1-NFE baselines. Furthermore, DBPO enables stable and effective policy improvement in online settings, and supports high-frequency real-world control.
\end{itemize}

%% file: secs/2_related.tex
\section{Related Work}
\label{sec:related}

\subsection{Diffusion-Based Visuomotor Policies and One-Step Acceleration}
Diffusion models have emerged as a central paradigm in robotic policy learning, as iterative denoising provides a flexible mechanism to represent multimodal action distributions. Diffusion Policy~\cite{Chi-RSS-23} formulates imitation learning as conditional denoising over action chunks, while DP3~\cite{Ze2024DP3} extends this formulation to 3D point-cloud observations. Subsequent variants enhance policy quality through architectural innovations and inductive-bias designs tailored for visuomotor control. A common limitation of this family of methods resides in deployment latency: each control query necessitates repeated denoising updates, thereby increasing closed-loop response times and interaction costs in online learning.

To mitigate these computational costs, recent studies investigate one-step acceleration via distillation and consistency-style objectives~\cite{prasad2024consistency,wang2024onestepdiffusionpolicyfast,salimans2022progressive,frans2024one,consistencypolicy2024,twostep2025,onedp2025,wang2025onedp,song2023improved}. These approaches can inherit robust behaviors from pretrained multi-step teacher models; however, the 1-NFE capability of these methods is typically acquired through compression pipelines or auxiliary acceleration stages. Consequently, one-step behavior frequently emerges as a result of post-hoc transformations rather than as an intrinsic property of the base policy objective.

\subsection{Flow-Style and Mean-Flow-Based One-Step Policy Learning}
In parallel, another prominent research direction pursues one-step generation via flow matching, consistency-style flow training, and mean-flow reformulations~\cite{lipman2023flow,song2023consistency,meanflow2024,flowmatching1,flowmatching2,rectifiedflow2022}. FlowPolicy, AdaFlow, and ManiFlow adapt flow-style policies for application in robotic manipulation~\cite{flowpolicy2024,hu2024adaflow,yan2025maniflow,zhang2024flowpolicy,maniflow2025}. Recent mean-flow-based methods, including MP1~\cite{sheng2025mp1}, DM1~\cite{zou2025dm1}, OMP~\cite{Fang2025OMPOM}, and MVP~\cite{zhan2026mvp}, demonstrate strong 1-NFE performance across challenging benchmarks~\cite{geng2025meanflow,improved_meanflow2025,splitmeanflow2025}.

In contrast to diffusion acceleration, these approaches avoid explicit teacher distillation and optimize one-step behavior in a more direct manner. However, they commonly rely on carefully designed auxiliary constraints---such as dispersive regularization, directional alignment, or instantaneous-velocity consistency---to stabilize the optimization process under strict one-step inference~\cite{diffuse_and_disperse}. This line of research demonstrates the practical feasibility of 1-NFE control, while simultaneously highlighting that the preservation of policy quality frequently depends on additional corrective objectives beyond the core generative mapping.

\subsection{Online RL for Generative Policies}
Recent studies further integrate online RL into generative policy backbones to elevate performance beyond the limits of offline imitation. DPPO~\cite{ren2025dppo} integrates PPO into multi-step diffusion policies, whereas ReinFlow~\cite{zhang2025reinflowfinetuningflowmatching}, DMPO~\cite{zou2026stepenoughdispersivemeanflow}, and MVP~\cite{zhan2026mvp} extend one-step flow and mean-flow frameworks via online optimization~\cite{d2ppo,liu2025flowgrpo,mcallister2025fmpg,lu2025vlarl}. Related efforts in domain-general generative RL similarly demonstrate that interaction-driven learning can enhance returns and robustness compared to static behavior cloning.

These advancements underscore a central trade-off in the online optimization of generative control: methods must achieve reward-driven adaptation from interaction while preserving deployment efficiency and policy stability under standard on-policy updates. The proposed method targets this regime by retaining strict 1-NFE execution and providing a compatible online optimization path.

%% file: secs/3_preliminaries.tex
\section{Preliminaries}
\label{sec:prelim}

This section briefly reviews the principle of Drifting Models (DM)~\cite{deng2026generative}. DM formulates generation as the training-time evolution of pushforward distributions; consequently, corrective dynamics are absorbed during optimization, and inference remains strictly one-step.

Let $\mathbf{z}\sim p_0=\mathcal{N}(\mathbf{0},\mathbf{I}_C)$ and let $f_{\theta}:\mathbb{R}^{C}\rightarrow\mathbb{R}^{D}$ denote a generator. A generated sample and its induced distribution are defined as:
\begin{equation}
\mathbf{x}=f_{\theta}(\mathbf{z}),
\qquad
q_{\theta}=[f_{\theta}]_{\#}p_0.
\label{eq:pre_sample}
\end{equation}
Here, $C$ and $D$ represent the latent and output dimensions, respectively, while $[f_{\theta}]_{\#}p_0$ denotes the pushforward of $p_0$ through $f_{\theta}$.

Let $p$ denote the target distribution on $\mathbb{R}^{D}$. During optimization, the parameters $\theta_k$ at iteration $k$ induce the sequence $q_k=[f_{\theta_k}]_{\#}p_0$. For a fixed latent seed, the sample $\mathbf{x}_k=f_{\theta_k}(\mathbf{z})$ evolves according to
\begin{equation}
\mathbf{x}_{k+1}
=
\mathbf{x}_k+\mathcal{V}_{p,q_k}(\mathbf{x}_k),
\label{eq:pre_drift_update}
\end{equation}
where $\mathcal{V}_{p,q}(\mathbf{x})\in\mathbb{R}^{D}$ denotes a drifting field.
DM adopts an anti-symmetric construction,
$\mathcal{V}_{p,q}(\mathbf{x})=-\mathcal{V}_{q,p}(\mathbf{x})$,
which implies that $\mathcal{V}_{p,p}(\mathbf{x})=\mathbf{0}$, thereby ensuring zero drift at distributional equilibrium.
This property motivates the formulation of a fixed-point target:
\begin{equation}
\tilde{\mathbf{x}}
=
\operatorname{sg}\!\left(
f_{\theta}(\mathbf{z})
+
\mathcal{V}_{p,q_{\theta}}\!\left(f_{\theta}(\mathbf{z})\right)
\right),
\label{eq:pre_drift_target}
\end{equation}
where $\operatorname{sg}(\cdot)$ represents the stop-gradient operation and $\tilde{\mathbf{x}}$ signifies the frozen drifted target.
The corresponding training objective is given by
\begin{equation}
\mathcal{L}_{\mathrm{DM}}
=
\mathbb{E}_{\mathbf{z}\sim p_0}
\left[
\left\|
f_{\theta}(\mathbf{z})-\tilde{\mathbf{x}}
\right\|_2^2
\right].
\label{eq:pre_drift_loss}
\end{equation}

Minimizing Eq.~\eqref{eq:pre_drift_loss} regresses the current predictions toward the drifted targets; consequently, field corrections are progressively encoded within the network parameters.
To obtain a computable field, DM employs a kernelized interaction formulation:
\begin{equation}
\mathcal{V}_{p,q}(\mathbf{x})
=
\mathbb{E}_{\mathbf{y}^{+}\sim p,\mathbf{y}^{-}\sim q}
\left[
\mathcal{K}(\mathbf{x},\mathbf{y}^{+},\mathbf{y}^{-})
\right],
\end{equation}
Here, $\mathbf{y}^{+}$ and $\mathbf{y}^{-}$ represent positive and negative samples drawn from $p$ and $q$, respectively, and
$\mathcal{K}:\mathbb{R}^{D}\times\mathbb{R}^{D}\times\mathbb{R}^{D}\rightarrow\mathbb{R}^{D}$ constitutes an interaction kernel.
In practice, this formulation yields an attraction toward target samples and a repulsion from model samples, while simultaneously preserving anti-symmetry.

In the proposed policy setting, generated samples correspond to action chunks, and the generator operates under observation conditioning. Section~\ref{sec:method} specializes this generic drifting objective for one-step robotic control and the online RL extension thereof.

%% file: secs/4_method.tex
\section{Method}
\label{sec:method}

This section introduces a two-stage framework that preserves single-pass deployment while enabling online policy improvement. Stage 1, DBP, learns a native one-step generative policy from offline demonstrations. Stage 2, DBPO, adds an exact-likelihood stochastic adapter for PPO-style updates while keeping strict 1-NFE execution. Figure~\ref{fig:method_framework} summarizes the pipeline. Additional method details are provided in the Supplementary Material.

Throughout this section, bold symbols denote vectors, matrices, or tensors. The index $t$ denotes environment step, $i$ minibatch sample, $r$ generated hypothesis index for a fixed observation, $h$ step index within an action chunk, $m$ scalar coordinate in $\mathbb{R}^{d_a}$, and $d$ flattened coordinate in $\mathbb{R}^{S}$.

\begin{figure*}[t]
    \centering
    \includegraphics[width=0.98\textwidth]{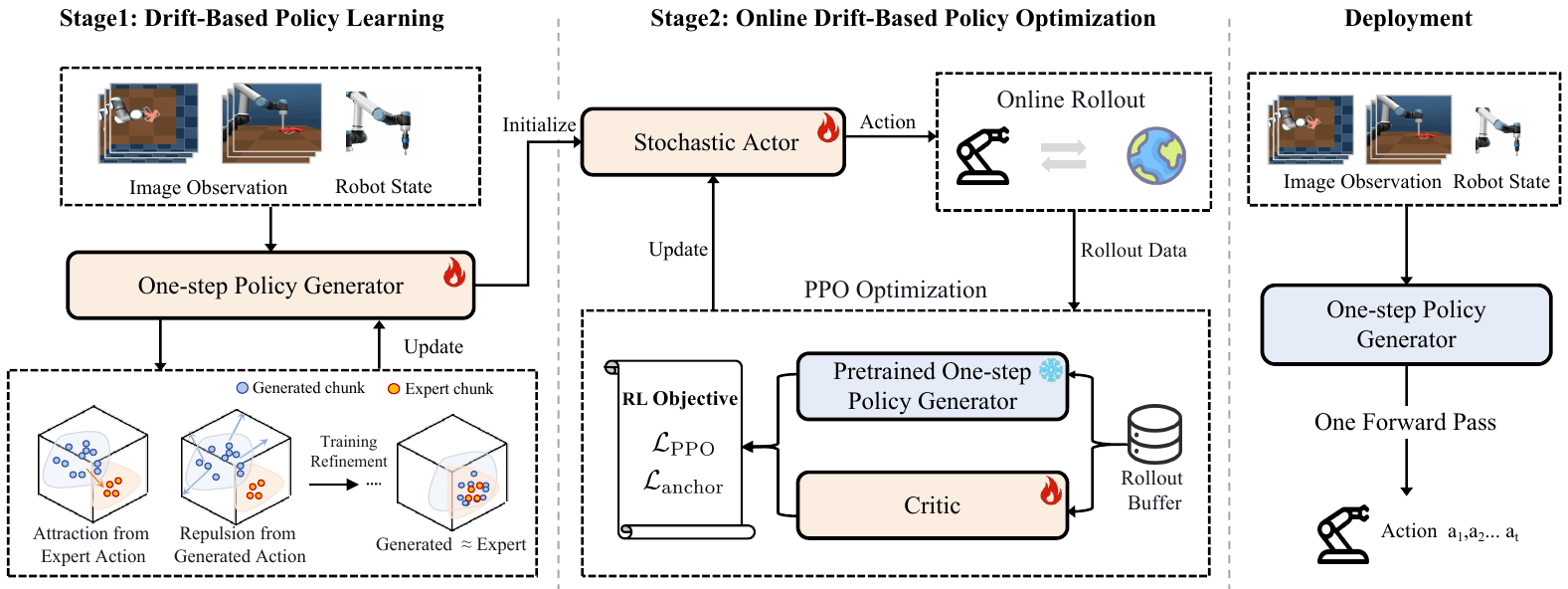}
    \caption{Two-stage Drift-Based Policy framework. Stage 1 learns a native one-step generator offline via attraction-repulsion refinement during training. Stage 2 fine-tunes a stochastic actor initialized from the pretrained backbone with on-policy PPO and anchor regularization, while deployment remains one-step (1-NFE).}
    \Description{A two-stage diagram showing offline Drift-Based Policy Learning (Stage 1) via attraction and repulsion, connected to Online Drift-Based Policy Optimization (Stage 2) using an exact-likelihood stochastic actor, rollout buffer, and a frozen reference for PPO regularization.}
    \label{fig:method_framework}
\end{figure*}

\subsection{Policy Setup}

At environment step $t$, let
$\mathbf{o}_t^{\mathrm{hist}}:=\mathbf{o}_{t-T_o+1:t}$
denote the observation history of length $T_o$. The policy predicts an action chunk of horizon $H$ in a single forward pass:
\begin{equation}
\mathbf{x}_t := [\mathbf{a}_t^{1},\ldots,\mathbf{a}_t^{H}] \in \mathbb{R}^{D},
\qquad
\mathbf{a}_t^{h} \in \mathbb{R}^{d_a},
\qquad
D = H d_a,
\label{eq:method_chunk}
\end{equation}
Here, $\mathbf{x}_t$ is the predicted chunk, $\mathbf{a}_t^{h}$ is the $h$-th action, $d_a$ is the per-step action dimension, and $D$ is the flattened chunk dimension.
Given an offline demonstration dataset
$\mathcal{D}=\{(\mathbf{o}_i^{\mathrm{hist}},\mathbf{x}_i^E)\}_{i=1}^{N}$,
where each pair contains an observation history and the aligned expert action chunk thereof, with
$\mathbf{x}_i^E=[\mathbf{a}_i^{E,1},\ldots,\mathbf{a}_i^{E,H}]$,
we use a one-step conditional generator with latent prior
$\mathbf{z}_t\sim p_0=\mathcal{N}(\mathbf{0},\mathbf{I})$ and fixed generation index $\tau=0$:
\begin{equation}
\begin{aligned}
\hat{\mathbf{x}}_t
&=
f_{\theta}(\mathbf{o}_t^{\mathrm{hist}},\mathbf{z}_t;\tau=0),\\
q_{\theta}(\cdot \mid \mathbf{o}_t^{\mathrm{hist}})
&=
[f_{\theta}(\mathbf{o}_t^{\mathrm{hist}},\cdot;\tau=0)]_{\#}p_0,
\end{aligned}
\label{eq:method_generator}
\end{equation}
Here, $f_{\theta}$ is the one-step generator, $\hat{\mathbf{x}}_t$ is the generated chunk, and $q_{\theta}(\cdot\mid\mathbf{o}_t^{\mathrm{hist}})$ is the induced conditional action distribution.
During deployment, receding-horizon control executes the sub-window starting at chunk index $T_o$ with an execution length $H_e$:
$\mathbf{x}_t^{\mathrm{exec}}=[\mathbf{a}_t^{T_o},\ldots,\mathbf{a}_t^{T_o+H_e-1}]$.
The boundary conditions $1\leq T_o\leq H$ and $1\leq H_e\leq H-T_o+1$ are strictly required to ensure the mathematical validity of the executed action slice.

\subsection{Drift-Based Policy Learning}

DBP trains $f_{\theta}$ by shaping a drift field in action space. Multiple hypotheses interact with expert anchors so that updates combine attraction to expert behavior and repulsion among hypotheses. The corrective dynamics are absorbed in training; deployment remains one-step.
For a trajectory minibatch $\{(\mathbf{o}_i^{\mathrm{hist}},\mathbf{x}_i^{E})\}_{i=1}^{B}$, we draw $G$ latent samples per observation and generate hypotheses
$\hat{\mathbf{x}}_i^{(r)}=f_{\theta}(\mathbf{o}_i^{\mathrm{hist}}, \mathbf{z}_i^{(r)};\tau=0)$, $r\in\{1,\ldots,G\}$.
Here, $G$ controls the number of hypotheses per observation.
During deployment, a single latent code is sampled and
$f_{\theta}(\mathbf{o}_t^{\mathrm{hist}}, \mathbf{z}_t;\tau=0)$ is evaluated once, precluding iterative refinement.

\noindent  \textbf{Drifting Objective in Action Space. }
Two training views are used. In chunk mode, trajectories are flattened with $S=Hd_a$ and drifting is applied once per sample. In step-wise mode, the same objective is applied to each step slice with $S=d_a$ and then averaged over $H$ steps.
For each minibatch, let $\mathbf{G}\in\mathbb{R}^{B\times G\times S}$ represent the generated hypotheses, let $\bar{\mathbf{G}}=\operatorname{sg}(\mathbf{G})$ be the detached copy thereof, and let $\mathbf{Y}=[\bar{\mathbf{G}},\mathbf{N}^{-},\mathbf{P}^{+}]$ denote the reference pool composed of generated references, optional negatives, and expert positives. Let $C_n=|\mathbf{N}^{-}|$, $C_p=|\mathbf{P}^{+}|$, and $U=G+C_n+C_p$. The negative-reference indices are denoted by $\mathcal{I}^{-}=\{1,\ldots,G+C_n\}$, and the positive-reference indices are denoted by $\mathcal{I}^{+}=\{G+C_n+1,\ldots,U\}$.
Omitting implementation constants, the core geometry is defined by
\begin{equation}
\begin{aligned}
d_{i,r,u}
&:=
\left\|\bar{\mathbf{G}}_{i,r,:}-\mathbf{Y}_{i,u,:}\right\|_2,\qquad
s_{\mathrm{norm}}
:=
\mathbb{E}_{i,r,u}[d_{i,r,u}],\\
A_{i,r,u}^{(R)}
&:=
\operatorname{SymSoftmax}\!\left(-\frac{d_{i,r,u}}{R\,s_{\mathrm{norm}}}\right), \qquad R\in\mathcal{R}.
\end{aligned}
\label{eq:method_scale_exact}
\end{equation}

Here, $d_{i,r,u}$ is the spatial pairwise distance, $s_{\mathrm{norm}}$ is the mean distance, and $A_{i,r,u}^{(R)}$ is the symmetric affinity at scale $R$. The set $\mathcal{R}$ contains finitely many interaction scales. A numerical floor is applied to ensure stability by $s_{\mathrm{norm}}>0$.

Define side masses
$S_{i,r,-}^{(R)}:=\sum_{u\in\mathcal{I}^{-}}A_{i,r,u}^{(R)}$ and
$S_{i,r,+}^{(R)}:=\sum_{u\in\mathcal{I}^{+}}A_{i,r,u}^{(R)}$.
The balanced coefficients are formulated as
\begin{equation}
\alpha_{i,r,u}^{(R)}
=
\begin{cases}
-A_{i,r,u}^{(R)}S_{i,r,+}^{(R)}, & u\in\mathcal{I}^{-},\\
\phantom{-}A_{i,r,u}^{(R)}S_{i,r,-}^{(R)}, & u\in\mathcal{I}^{+}.
\end{cases}
\label{eq:method_alpha_exact}
\end{equation}

This formulation yields repulsion from $\mathcal{I}^{-}$ and attraction toward $\mathcal{I}^{+}$ with cross-side mass balancing. In particular,
$\sum_{u\in\mathcal{I}^{-}}\alpha_{i,r,u}^{(R)}=-\sum_{u\in\mathcal{I}^{+}}\alpha_{i,r,u}^{(R)}$,
which enforces antisymmetric mass exchange between the two sides.

For each scale $R$, the drift contribution is
$\mathbf{F}_{i,r,:}^{(R)}:=\sum_{u=1}^{U}\alpha_{i,r,u}^{(R)}(\mathbf{Y}_{i,u,:}-\bar{\mathbf{G}}_{i,r,:})/s_{\mathrm{norm}}$.
After per-scale RMS normalization and aggregation,
$\mathbf{V}_{i,r,:}:=\sum_{R\in\mathcal{R}}\widehat{\mathbf{F}}_{i,r,:}^{(R)}$.
The fixed-point regression form is
% \begin{equation}
% \tilde{\mathbf{X}}
% =
% \operatorname{sg}\!\left(\bar{\mathbf{G}}/s_{\mathrm{norm}}+\mathbf{V}\right),
% \qquad
% \ell_i
% =
% \frac{1}{GS}\sum_{r=1}^{G}\sum_{d=1}^{S}
% \left(\frac{G_{i,r,d}}{s_{\mathrm{norm}}}-\tilde{X}_{i,r,d}\right)^2,
% \label{eq:method_target_loss}
% \end{equation}
\begin{equation}
\begin{aligned}
\tilde{\mathbf{X}}
&=
\operatorname{sg}\!\left(\bar{\mathbf{G}}/s_{\mathrm{norm}}+\mathbf{V}\right), \\
\ell_i
&=
\frac{1}{GS}\sum_{r=1}^{G}\sum_{d=1}^{S}
\left(\frac{G_{i,r,d}}{s_{\mathrm{norm}}}-\tilde{X}_{i,r,d}\right)^2,
\end{aligned}
\label{eq:method_target_loss}
\end{equation}
where $\operatorname{sg}(\cdot)$ is stop-gradient and $\mathbf{V}$ is the aggregated multi-scale drift.
The final DBP objective constitutes the sample average in chunk mode and the time-averaged sample loss in step-wise mode:
\begin{equation}
\mathcal{L}_{\mathrm{DBP}}
=
\begin{cases}
\frac{1}{B}\sum_{i=1}^{B}\ell_{i}, & \text{chunk mode},\\
\frac{1}{BH}\sum_{h=1}^{H}\sum_{i=1}^{B}\ell_{i}^{(h)}, & \text{step-wise mode},
\end{cases}
\label{eq:method_ldbp}
\end{equation}
where $\ell_{i}^{(h)}$ is the sample loss on the $h$-th action slice with $S=d_a$.
Detailed pseudocode, numerical stabilizers, and additional derivations are provided in the Supplementary Material for readability.

\noindent \textbf{Principled Divergence View and Convergence Guarantee.}
From Eq.~\eqref{eq:method_target_loss}, the chunk-mode objective equals the mean squared drift magnitude,
$\mathcal{L}_{\mathrm{DBP}}=\mathcal{D}_{\mathrm{drift}}(q_{\theta},p;\mathcal{R})$,
because $\bar{\mathbf{G}}=\operatorname{sg}(\mathbf{G})$ keeps forward values unchanged while blocking gradients in target construction. Here, $p$ is the expert conditional action distribution and $q_{\theta}$ is the generated conditional distribution.

Under standard stochastic approximation assumptions (smooth $\mathcal{D}_{\mathrm{drift}}$, unbiased bounded-variance minibatch gradients, and Robbins--Monro step sizes $\sum_k\eta_k=\infty$, $\sum_k\eta_k^2<\infty$), SGD on Eq.~\eqref{eq:method_ldbp} reaches a first-order stationary regime, i.e.,
$\liminf_{k\to\infty}\mathbb{E}[\|\nabla_{\theta}\mathcal{D}_{\mathrm{drift}}(\theta_k)\|_2^2]=0$.
If, additionally, the generator Jacobian is locally full-rank on the support of $p_0$ and the drifting field is identifiable at equilibrium (i.e., $\mathcal{V}_{p,q}(\mathbf{x})\equiv\mathbf{0}\Rightarrow q=p$), zero-drift stable points correspond to the target distribution in the induced action space. This identifiability statement is used as an additional sufficient assumption.

Numerical stabilizers (e.g., clipping, masking constants, and finite-precision safeguards) constitute implementation details and do not alter the aforementioned optimization principle.

\subsection{Online Drift-Based Policy Optimization}

Although DBP provides a robust one-step offline initialization, offline imitation alone is insufficient to address reward-driven distribution shifts encountered during online interaction. The extension of DBP to online RL must concurrently satisfy three requirements: preserving the pretrained one-step generator, providing exact likelihoods for standard PPO~\cite{schulman2017proximal}, and maintaining deployment at 1-NFE. DBPO fulfills these requirements via a minimal stochastic adapter constructed upon the pretrained backbone.

For brevity within this subsection, let $\mathbf{o}_t:=\mathbf{o}_t^{\mathrm{hist}}$. In Eq.~\eqref{eq:method_generator}, the marginal $q_{\theta}(\cdot\mid\mathbf{o}_t)$ is implicit because it integrates over latent variables, whereas PPO requires explicit rollout likelihoods. DBPO therefore uses a trainable active backbone, a frozen reference backbone, and an analytic stochastic actor $\pi_{\theta,\psi}$.

\noindent \textbf{Exact-Likelihood Stochastic Actor. }
For generic inputs $(\mathbf{o},\mathbf{z})$, the active backbone predicts the latent-conditioned chunk mean $\boldsymbol{\mu}_{\theta}(\mathbf{o},\mathbf{z})$ and an observation feature representation $\mathbf{c}_{\theta}(\mathbf{o})$. A diagonal log-standard-deviation head is attached as
$\log\boldsymbol{\sigma}_{\psi}(\mathbf{o}) := g_{\psi}(\mathbf{c}_{\theta}(\mathbf{o}))$, with
$\log\boldsymbol{\sigma}_{\psi}(\mathbf{o})\in\mathbb{R}^{D}$ spanning the action dimension.

The resulting actor is defined as:
\begin{equation}
\pi_{\theta,\psi}(\mathbf{x}\mid\mathbf{o},\mathbf{z})
=
\mathcal{N}\!\left(
\mathbf{x};
\boldsymbol{\mu}_{\theta}(\mathbf{o},\mathbf{z}),
\operatorname{diag}\!\left(\boldsymbol{\sigma}_{\psi}(\mathbf{o})^2\right)
\right).
\label{eq:method_actor}
\end{equation}
Here, $\boldsymbol{\sigma}_{\psi}(\mathbf{o})$ controls exploration noise. During rollout, the policy samples $\mathbf{z}_t\sim p_0$ and then $\mathbf{x}_t\sim\pi_{\theta,\psi}(\cdot\mid\mathbf{o}_t,\mathbf{z}_t)$, and stores $(\mathbf{z}_t,\mathbf{x}_t)$ in the buffer. Because PPO reuses the same stored latent $\mathbf{z}_t$ in evaluation epochs, Eq.~\eqref{eq:method_actor} provides exact conditional rollout likelihoods.

\noindent \textbf{Executed-Prefix Likelihood. }
Online updates are restricted to the executed prefix
$\mathbf{x}_t^{\mathrm{exec}}=[\mathbf{a}_t^{T_o},\ldots,\mathbf{a}_t^{T_o+H_e-1}]$.
Its conditional log-likelihood is
% \begin{multline}
% \log\pi_{\theta,\psi}\!\left( \mathbf{x}_t^{\mathrm{exec}}\mid\mathbf{o}_t,\mathbf{z}_t \right) \\
% = \sum_{h=T_o}^{T_o+H_e-1}\sum_{m=1}^{d_a}
% \log\mathcal{N}\!\Big(a_{t,m}^{h};\,\mu_{\theta,m}^{h}(\mathbf{o}_t,\mathbf{z}_t),\,\sigma_{\psi,m}^{h}(\mathbf{o}_t)^2\Big).
% \label{eq:method_logprob}
% \end{multline}
\begin{equation}
\begin{aligned}
\log\pi_{\theta,\psi}\!\left( \mathbf{x}_t^{\mathrm{exec}}\mid\mathbf{o}_t,\mathbf{z}_t \right)
&= \\ \sum_{h=T_o}^{T_o+H_e-1}\sum_{m=1}^{d_a}
\log\mathcal{N}\!\Big(
a_{t,m}^{h}, 
&\quad
\mu_{\theta,m}^{h}(\mathbf{o}_t,\mathbf{z}_t),\,
\sigma_{\psi,m}^{h}(\mathbf{o}_t)^2
\Big).
\end{aligned}
\label{eq:method_logprob}
\end{equation}
Here, $h\in\{T_o,\ldots,T_o+H_e-1\}$ indexes executed steps and $m\in\{1,\ldots,d_a\}$ indexes scalar action coordinates. The terms $\mu_{\theta,m}^{h}(\mathbf{o}_t,\mathbf{z}_t)$ and $\sigma_{\psi,m}^{h}(\mathbf{o}_t)$ are scalar components of the actor mean and standard deviation in Eq.~\eqref{eq:method_actor}.
Under closed-loop re-planning, unexecuted suffix actions are replaced at the next environment step and therefore do not enter current-step credit assignment. Consequently, optimizing only the executed prefix remains consistent with on-policy rollouts used to estimate $\hat{A}_t$.

\noindent \textbf{Joint-Policy View and Ratio Equivalence. }
Define the joint policy as $\tilde{\pi}_{\theta,\psi}(\mathbf{x}^{\mathrm{exec}},\mathbf{z}\mid\mathbf{o}) := p_0(\mathbf{z})\,\pi_{\theta,\psi}(\mathbf{x}^{\mathrm{exec}}\mid\mathbf{o},\mathbf{z})$.
Here, $p_0(\mathbf{z})$ is fixed and independent of $(\theta,\psi)$. Using the same stored latent $\mathbf{z}_t$ for new and old policies, the PPO importance ratio becomes
\begin{equation}
\begin{aligned}
\tilde{r}_t(\theta,\psi)
&:=
\frac{\tilde{\pi}_{\theta,\psi}(\mathbf{x}_t^{\mathrm{exec}},\mathbf{z}_t\mid\mathbf{o}_t)}{\tilde{\pi}_{k}(\mathbf{x}_t^{\mathrm{exec}},\mathbf{z}_t\mid\mathbf{o}_t)}
=
\frac{\pi_{\theta,\psi}(\mathbf{x}_t^{\mathrm{exec}}\mid\mathbf{o}_t,\mathbf{z}_t)}{\pi_{k}(\mathbf{x}_t^{\mathrm{exec}}\mid\mathbf{o}_t,\mathbf{z}_t)}
=: r_t(\theta,\psi),
\end{aligned}
\label{eq:method_ratio_equiv}
\end{equation}
which demonstrates the exact equivalence between the joint-policy PPO ratio and the conditional ratio employed in DBPO. Consequently, computationally expensive marginalization over the latent variable $\mathbf{z}$ is not required during policy optimization.

\noindent \textbf{PPO Objective with Drift-Based Anchor. }
Let $\pi_k$ denote the behavior policy responsible for rollout collection, and let $\hat{A}_t$ denote the advantage estimate. Following standard PPO~\cite{schulman2017proximal}, the importance ratio $r_t(\theta,\psi)$ from Eq.~\eqref{eq:method_ratio_equiv} is utilized to represent the PPO objective compactly as
\begin{equation}
\mathcal{L}_{\mathrm{PPO}}
=
\mathcal{L}_{\mathrm{clip}}(r_t,\hat{A}_t)
+
c_v\mathcal{L}_{\mathrm{value}}
-
c_e\mathcal{H},
\label{eq:method_lppo_compact}
\end{equation}
where $\mathcal{L}_{\mathrm{clip}}$ is the clipped surrogate, $\mathcal{L}_{\mathrm{value}}$ is value regression, $\mathcal{H}$ is the entropy bonus, and $c_v,c_e>0$ are weights. To reduce drift from the pretrained state $\bar{\theta}$, we use the anchor loss:
\begin{equation}
\mathcal{L}_{\mathrm{anchor}}
=
\mathbb{E}_t\!\left[
\left\|
\boldsymbol{\mu}_{\theta}(\mathbf{o}_t,\mathbf{z}_t)
-
\boldsymbol{\mu}_{\bar{\theta}}(\mathbf{o}_t,\mathbf{z}_t)
\right\|_2^2
\right].
\label{eq:method_lanchor}
\end{equation}

This term penalizes the distance between updated and frozen mean predictions under identical latent inputs, which stabilizes policy updates around the pretrained operating region. The complete objective is formulated as:
\begin{equation}
\mathcal{L}_{\mathrm{RL}}
=
\mathcal{L}_{\mathrm{PPO}}
+
\lambda_{\mathrm{anchor}}\mathcal{L}_{\mathrm{anchor}}.
\label{eq:method_lrl}
\end{equation}
Here, $\lambda_{\mathrm{anchor}}$ acts as the scalar weight for anchor regularization.

During training, exploration operates via Gaussian noise superimposed on the mean, whereas deployment removes this noise and reverts to the deterministic center. Throughout closed-loop manipulation, the execution of a prediction necessitates exactly one network evaluation, thereby maintaining 1-NFE efficiency.

%% file: secs/5_experiments.tex
\section{Experiments}
\label{sec:exp}

The experiments evaluate three primary aspects: (1) the capability of the proposed Drift-Based Policy (DBP) to match or exceed the iterative diffusion baseline while reducing inference to exactly 1-NFE; (2) the robustness of the DBP backbone on large-scale 3D point-cloud manipulation benchmarks; and (3) the cross-domain performance gains and real-robot transfer efficacy of its online reinforcement learning extension, Drift-Based Policy Optimization (DBPO).

\subsection{Evaluation Setup and Protocols}

\textbf{Dataset.}
We evaluate on three benchmark families. (i) In the reproduced Diffusion Policy suite~\cite{Chi-RSS-23}, we use Push-T (Image), Push-T (Low-Dim), BlockPush (P1/P2), RoboMimic (Low-Dim), RoboMimic (Image), and Kitchen (12 tasks in total). (ii) For point-cloud one-step evaluation, we follow the MP1/OMP protocol on 37 tasks: 3 Adroit tasks and 34 Meta-World tasks (21 Easy, 4 Medium, 4 Hard, and 5 Very Hard)~\cite{Rajeswaran-RSS-18,mclean2025metaworld,sheng2025mp1,Fang2025OMPOM}. (iii) For online RL, we evaluate on 4 RoboMimic manipulation tasks and D4RL locomotion tasks, using the same simplified RoboMimic setting as prior one-step RL baselines~\cite{robomimic2021,d4rl2020,ren2025dppo,zhang2025reinflowfinetuningflowmatching,zou2026stepenoughdispersivemeanflow}.

\noindent \textbf{Metrics. }
For manipulation tasks, we report success rate; for D4RL locomotion, we report episode return. Computational efficiency is measured by NFE. In the reproduced Diffusion Policy suite, each result is averaged over the last 10 checkpoints, and each checkpoint is averaged over 3 training seeds. In the point-cloud protocol, evaluation is performed every 200 epochs, the top-5 checkpoints per seed are averaged, and mean$\pm$std across seeds is reported. For online RL benchmarks, each task is evaluated with 100 episodes.

\noindent \textbf{Baselines. }
We explicitly separate reproduced and quoted baselines. Reproduced baselines include Diffusion Policy in the diffusion suite, and ReinFlow~\cite{zhang2025reinflowfinetuningflowmatching} plus DMPO~\cite{zou2026stepenoughdispersivemeanflow} in online RL comparisons. For Adroit/Meta-World point-cloud comparisons, non-ours results are quoted from OMP~\cite{Fang2025OMPOM}. Our method follows the same MP1/OMP architecture and training protocol, including the same U-Net backbone, for protocol-matched comparison.

\noindent \textbf{Implementation Details. }
To isolate algorithmic effects from architectural effects, we match baseline architectures whenever reproduction is performed. In the point-cloud setting, we use the same U-Net backbone and training pipeline as MP1/OMP, without stronger backbones or additional data augmentation. We use 10 demonstrations per task, FPS preprocessing to 512/1024 points, and $84\times84$ image resizing when applicable. Seeds are \{0,1,2\}. Training runs for 3000 epochs on Adroit and 1000 epochs on Meta-World on 8 $\times$ NVIDIA RTX 3090 GPUs. Unless otherwise specified, the default temperature set is $\mathcal{R}=\{0.02,0.05,0.2\}$. Additional tuning shows task-dependent optima, and task-specific temperature sets are used when dedicated tuning is reported.

\subsection{DBP Compared with Multi-Step Policy}

We first test whether DBP preserves policy quality when iterative denoising is replaced by strict 1-NFE inference under the reproduced Diffusion Policy setting.

\begin{figure}[t]
\centering
\includegraphics[width=\linewidth]{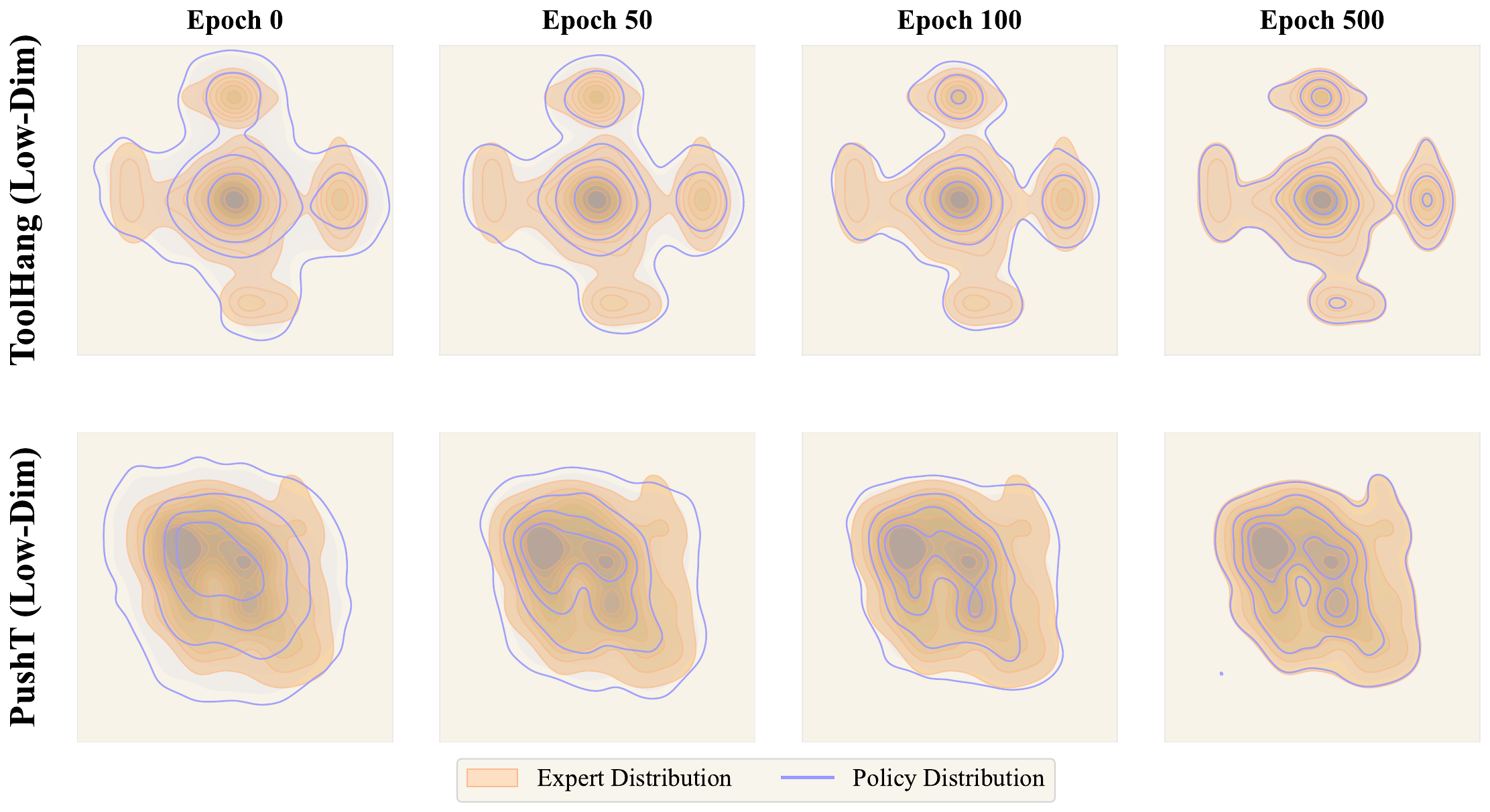}
\caption{Evolution of the internalized drift manifold. The policy action distribution (blue) progressively aligns with expert modes (peach) during training.}
\Description{A sequence of action-distribution plots in which blue policy samples progressively contract toward the peach expert modes over training.}
\label{fig:mode_alignment_combined}
\end{figure} 

\input{tabs/tab_dp_summary_main}

\input{tabs/tab_pointcloud_main}

Table~\ref{tab:dp_summary_main} summarizes the reproduced comparison. DBP increases the task average from 0.79 to 0.83 while reducing inference from 100 NFE to 1 NFE, corresponding to a 100$\times$ speedup. At the task level, DBP improves Push-T (Low-Dim), BlockPush, and RoboMimic (Low-Dim), ties Kitchen, and is slightly lower than Diffusion Policy on Push-T (Image) and RoboMimic (Image). These results show that drifting can absorb most iterative correction effects into a one-step mapping while retaining substantial efficiency gains.

Figure~\ref{fig:mode_alignment_combined} visualizes this transition. Early in training, generated actions are dispersed relative to expert modes; as optimization proceeds, the distribution contracts toward expert-supported regions. This trend is consistent with the drifting hypothesis that corrective dynamics are internalized during training and executed by a single forward pass at deployment.

\subsection{DBP Compared with One-Step Baselines}

We evaluate the native one-step quality of DBP before online adaptation, with the goal of isolating the offline backbone under strict 1-NFE inference. Following the MP1/OMP protocol, we use matched backbone and training settings; non-ours results are quoted from OMP for protocol-consistent comparison.

As shown in Table~\ref{tab:pointcloud_main}, DBP achieves the best 37-task average success rate of 88.4\%$\pm$3.1, outperforming OMP (82.3\%) and MP1 (78.9\%) while preserving one-step deployment. This result shows that the proposed native one-step backbone maintains strong policy quality at scale.

The gains are broad across difficulty groups: relative to OMP, DBP improves Easy by +2.0, Medium by +12.9, Hard by +12.7, and Very Hard by +8.9. On the dexterous manipulation benchmark Adroit, DBP matches the best Hammer result, improves Door, and delivers a substantial gain on Pen (+20.0 over OMP). Compared with diffusion-style baselines in the same table (DP/DP3 at 10 NFE), DBP also attains higher overall performance under a stricter one-step inference budget. These results support robust one-step manipulation performance and provide a stronger offline initialization for subsequent online fine-tuning.

\subsection{Online Fine-Tuning of the DBP Backbone}

We next evaluate whether PPO fine-tuning improves the pretrained drift-based one-step backbone and whether such gains transfer from sparse-reward manipulation to locomotion.

\begin{figure}[htbp]
    \centering
    \begin{subfigure}{\linewidth}
        \includegraphics[width=\linewidth]{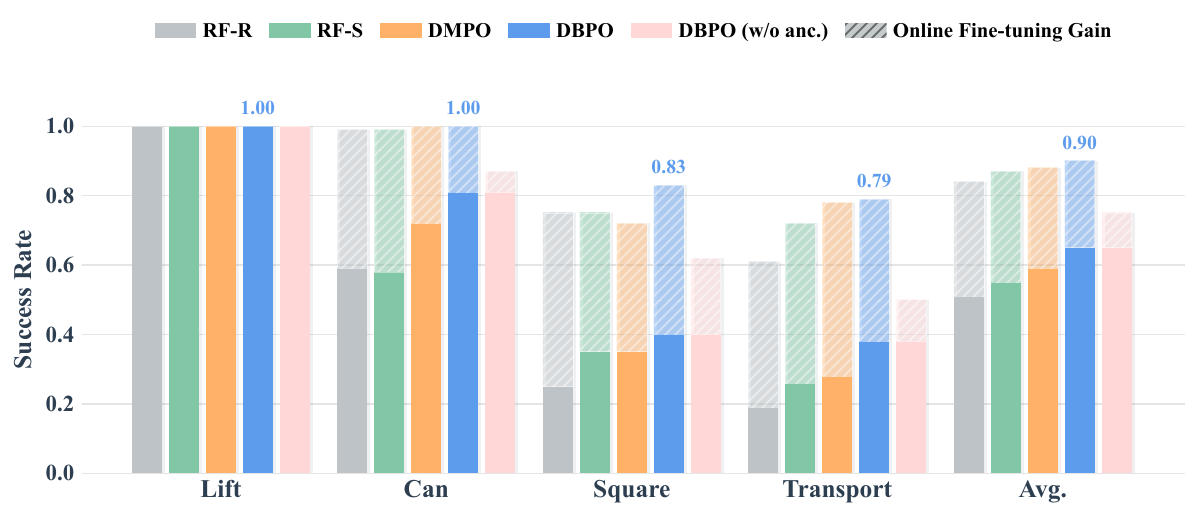}
        \caption{RoboMimic (image-based): offline initialization and PPO fine-tuning gains.}
        \label{fig:robomimic_bars}
    \end{subfigure}
    
    \begin{subfigure}{\linewidth}
        \includegraphics[width=\linewidth]{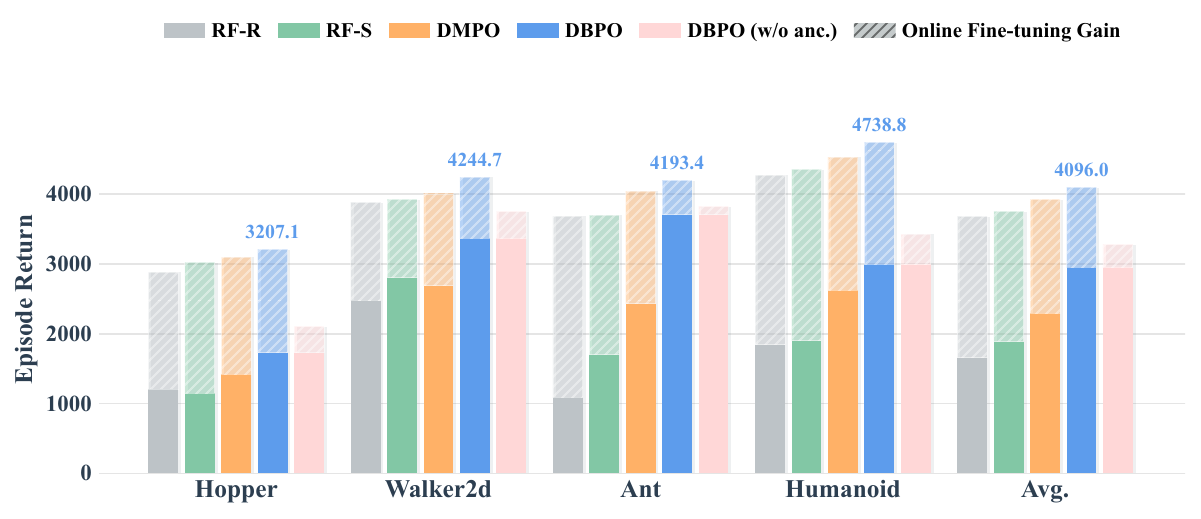}
        \caption{D4RL locomotion: offline initialization and PPO fine-tuning gains.}
        \label{fig:d4rl_bars}
    \end{subfigure}
    
    \caption{Online PPO fine-tuning results on RoboMimic and D4RL with anchor ablation (DBPO vs. DBPO w/o anchor). Solid bars denote offline initialization, and hatched bars denote gains after fine-tuning. DBPO achieves the strongest post-fine-tuning performance, while removing the anchor reduces gains over pretrained baselines.}
    \Description{Two bar charts comparing offline initialization and PPO fine-tuning gains on RoboMimic and D4RL. DBPO has the strongest final performance, and its no-anchor ablation gains less.}
    \label{fig:combined_online_results}
\end{figure} 

\begin{figure*}[t]
    \centering
    \includegraphics[width=\textwidth]{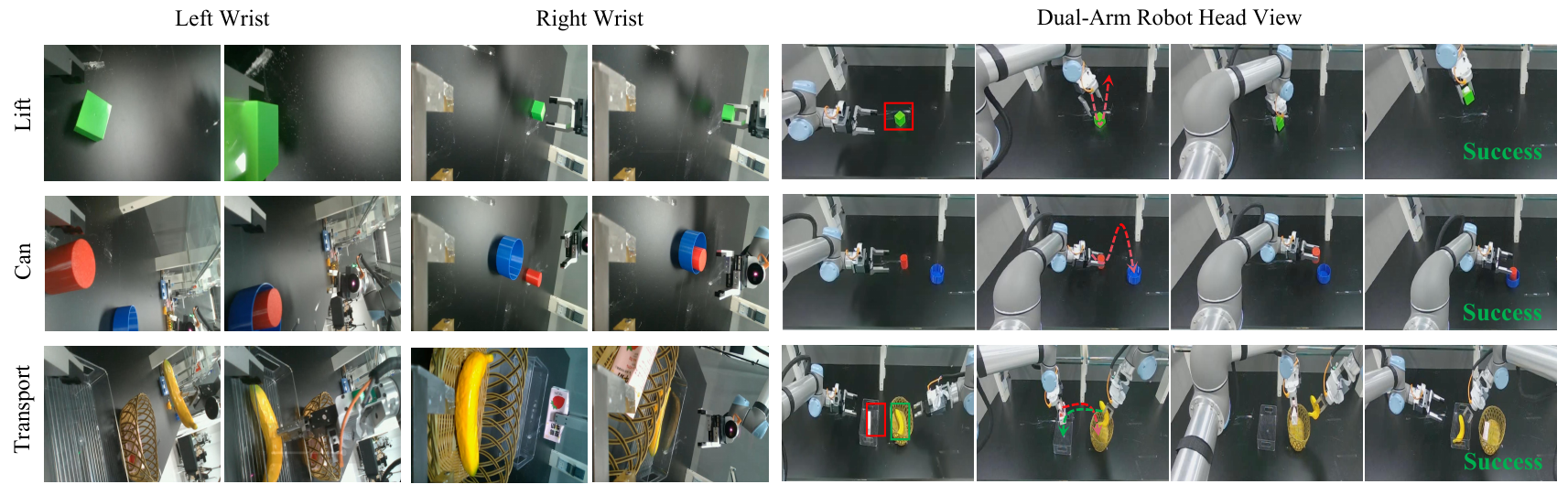}
    \caption{Real-world bimanual deployment on the physical UR5 testbed. Drift-Based Policy executes precision Lift, Can, and synchronized bimanual Transport using raw trilateral camera inputs.}
    \Description{A real-world deployment figure showing successive frames of executions by Drift-Based Policy on Lift, Can, and Transport tasks from head and dual-wrist camera angles.} 
    \label{fig:realrobot}
\end{figure*} 

\noindent \textbf{RoboMimic Policy Learning and PPO Fine-Tuning. }
We compare DBPO with reproduced one-step RL baselines (ReinFlow and DMPO) on sparse-reward, image-based RoboMimic tasks. Figure~\ref{fig:robomimic_bars} separates offline initialization (solid bars) from fine-tuning gains (hatched bars), enabling a direct comparison of backbone quality and online adaptation effectiveness. DBPO starts from a stronger one-step initialization and further improves after PPO fine-tuning while preserving 1-NFE deployment.

\noindent \textbf{D4RL Gym Locomotion. }
To test cross-domain transfer, we apply the same backbone and online adapter to D4RL locomotion tasks. Figure~\ref{fig:d4rl_bars} shows that DBPO attains the highest average return among compared native one-step RL baselines. Together with the RoboMimic results, this shows robust transfer across domains with different reward structures and action dynamics.

\noindent \textbf{Anchor Ablation in Online Fine-Tuning. }
To validate the anchor regularizer, we evaluate a variant without it (DBPO w/o anchor) on RoboMimic and D4RL. Starting from the same pretrained initialization, full DBPO achieves average scores of 0.90 on RoboMimic and 4096.0 on D4RL (Figure~\ref{fig:combined_online_results}). In contrast, removing the anchor leads to performance drops to 0.75 and 3273.5, respectively. This consistent degradation confirms that the anchor is crucial for mitigating representation drift and restricting arbitrary policy deviation from the pretrained prior, thereby effectively stabilizing the online fine-tuning process.

\subsection{Real-World Deployment}

\noindent \textbf{Experimental Setup and Results. }We deploy DBP on a physical dual-arm UR5 setup with an NVIDIA RTX 3090 GPU and a tri-camera array (dual wrist-mounted RealSense L515 and one Orbbec Gemini head camera). We collect 50 teleoperation demonstrations per task and evaluate the policy on RoboMimic-style real-world Lift, Can, and synchronized bimanual Transport tasks~\cite{robomimic2021,aloha}. Figure~\ref{fig:realrobot} shows representative execution traces. Table~\ref{tab:real_robot_summary} reports 50 trials per task and includes a direct comparison against the one-step baseline MP1 on the same platform. DBP reaches 123/150 successes (82\%), compared with 89/150 (59\%) for MP1.

Under visual occlusion and hardware constraints, DBP maintains reliable execution without any modification to the 1-NFE computation path. The average end-to-end latency is 9.5~ms, supporting real-time closed-loop control.

\noindent \textbf{Failure Modes and System Integration. }Failures come from two recurring cases: grasp slip in Can and inter-arm action conflict in Transport. During deployment, DBP runs as the high-level action generator, while the low-level dual-arm control pipeline remains unchanged. No modification to the underlying hardware interface or control stack is required.

\input{tabs/tab_real_robot}

%% file: tabs/tab_dp_summary_main.tex
\begin{table}[tb]
\centering
\scriptsize
\setlength{\tabcolsep}{3pt}
\renewcommand{\arraystretch}{0.95}
\caption{Comparison between Diffusion Policy and Ours on the Diffusion Policy suite. BlockPush/Kitchen are phase-averaged. Entries report success rate. Best results are in \textbf{bold}.}
\label{tab:dp_summary_main}
\resizebox{0.96\columnwidth}{!}{
\begin{tabular}{@{}lcc@{}}
\toprule
Task & Diffusion Policy (100 NFE) & Ours (1 NFE) \\
\midrule
Push-T (Image) & \textbf{0.91} & 0.89 \\
Push-T (Low-Dim) & 0.85 & \textbf{0.87} \\
BlockPush (P1/P2) & 0.24 & \textbf{0.43} \\
RoboMimic (Low-Dim) & 0.80 & \textbf{0.92} \\
RoboMimic (Image) & \textbf{0.91} & 0.87 \\
Kitchen (P1/P2/P3/P4) & \textbf{1.00} & \textbf{1.00} \\
\midrule
\rowcolor[gray]{0.92} Avg. & 0.79 & \textbf{0.83} \\
\bottomrule
\end{tabular}
}
\renewcommand{\arraystretch}{1}
\end{table}

%% file: tabs/tab_pointcloud_main.tex
\begin{table*}[t]
\centering
\scriptsize
\setlength{\tabcolsep}{3pt}
\renewcommand{\arraystretch}{0.95}
\caption{Point-cloud imitation comparison on Adroit and Meta-World under the MP1/OMP protocol. Ours refers to Drift-Based Policy. Entries report success rate (\%, mean $\pm$ std over 3 seeds). Best results are in \textbf{bold}; second-best are \underline{underlined}.}
\label{tab:pointcloud_main}
\resizebox{0.96\textwidth}{!}{
\begin{tabular}{@{}lccccccccc@{}}
\toprule
& & \multicolumn{3}{c}{Adroit} & \multicolumn{4}{c}{Meta-World} & \\
\cmidrule(lr){3-5}\cmidrule(lr){6-9}
Method & NFE & Hammer & Door & Pen & Easy (21) & Medium (4) & Hard (4) & Very Hard (5) & Average \\
\midrule
DP & 10 & 16.0$\pm$10.0 & 34.0$\pm$11.0 & 13.0$\pm$2.0 & 50.7$\pm$6.1 & 11.0$\pm$2.5 & 5.25$\pm$2.5 & 22.0$\pm$5.0 & 35.2$\pm$5.3 \\
DP3 & 10 & \textbf{100.0}$\pm$0.0 & 56.0$\pm$5.0 & 46.0$\pm$10.0 & 87.3$\pm$2.2 & 44.5$\pm$8.7 & 32.7$\pm$7.7 & 39.4$\pm$9.0 & 68.7$\pm$4.7 \\
Simple DP3 & 10 & 98.0$\pm$2.0 & 40.0$\pm$17.0 & 36.0$\pm$4.0 & 86.8$\pm$2.3 & 42.0$\pm$6.5 & 38.7$\pm$7.5 & 35.0$\pm$11.6 & 67.4$\pm$5.0 \\
\midrule
Adaflow & --- & 45.0$\pm$11.0 & 27.0$\pm$6.0 & 18.0$\pm$6.0 & 49.4$\pm$6.8 & 12.0$\pm$5.0 & 5.75$\pm$4.0 & 24.0$\pm$4.8 & 35.6$\pm$6.1 \\
CP & 1 & 45.0$\pm$4.0 & 31.0$\pm$10.0 & 13.0$\pm$6.0 & 69.3$\pm$4.2 & 21.2$\pm$6.0 & 17.5$\pm$3.9 & 30.0$\pm$4.9 & 50.1$\pm$4.7 \\
FlowPolicy & 1 & 98.0$\pm$1.0 & 61.0$\pm$2.0 & 54.0$\pm$4.0 & 84.8$\pm$2.2 & 58.2$\pm$7.9 & 40.2$\pm$4.5 & 52.2$\pm$5.0 & 71.6$\pm$3.5 \\
MP1 & 1 & \textbf{100.0}$\pm$0.0 & \underline{69.0}$\pm$2.0 & 58.0$\pm$5.0 & 88.2$\pm$1.1 & 68.0$\pm$3.1 & 58.1$\pm$5.0 & 67.2$\pm$2.7 & 78.9$\pm$2.1 \\
OMP & 1 & \textbf{100.0}$\pm$0.0 & 68.0$\pm$3.0 & \underline{60.0}$\pm$4.0 & \underline{89.7}$\pm$0.7 & \underline{77.4}$\pm$2.2 & \underline{62.5}$\pm$3.1 & \underline{77.8}$\pm$3.0 & \underline{82.3}$\pm$1.6 \\
\midrule
\rowcolor[gray]{0.92} Ours & 1 & \textbf{100.0}$\pm$0.0 & \textbf{70.0}$\pm$2.0 & \textbf{80.0}$\pm$6.0 & \textbf{91.7}$\pm$1.7 & \textbf{90.3}$\pm$3.6 & \textbf{75.2}$\pm$6.1 & \textbf{86.7}$\pm$5.8 & \textbf{88.4}$\pm$3.1 \\
\bottomrule
\end{tabular}
}
\renewcommand{\arraystretch}{1}
\end{table*}

%% file: tabs/tab_real_robot.tex
\begin{table}[htbp]
\centering
\scriptsize
\setlength{\tabcolsep}{10pt}
\caption{Additional real-robot dual-arm deployment results on the physical UR5 setup (50 trials per task).}
\label{tab:real_robot_summary}
\resizebox{\columnwidth}{!}{
\begin{tabular}{lcc}
\toprule
\textbf{Task} & \textbf{DBP (Ours)} & \textbf{MP1 (one-step baseline)} \\
\midrule
Lift (single-arm)      & \textbf{47/50 (94\%)} & 38/50 (76\%) \\
Can (single-arm)       & \textbf{43/50 (86\%)} & 32/50 (64\%) \\
Transport (dual-arm)   & \textbf{33/50 (66\%)} & 19/50 (38\%) \\
\midrule
\rowcolor[gray]{0.92} \textbf{Overall} & \textbf{123/150 (82\%)} & 89/150 (59\%) \\
\bottomrule
\end{tabular}
}
\end{table}

%% file: secs/6_conclusion.tex
\section{Conclusion}
\label{sec:conclusion}

We presented Drift-Based Policy (DBP), a native one-step generative policy that transfers iterative refinement from inference to fixed-point training, and Drift-Based Policy Optimization (DBPO), which enables PPO-style online fine-tuning through exact conditional rollout likelihoods while preserving 1-NFE deployment. DBP improves the average success rate from 0.79 to 0.83 on the 12-task Diffusion Policy suite while reducing inference from 100 NFEs to 1, and achieves an average success rate of 88.4\% across 37 point-cloud manipulation tasks, surpassing the leading 1-NFE baseline at 82.3\%. DBPO further improves the pretrained policy on RoboMimic and D4RL, with anchor regularization stabilizing online updates. On a physical dual-arm UR5 platform, DBP achieves 123/150 successes (82\%) with an average end-to-end latency of 9.5~ms, compared with MP1's 89/150 successes (59\%). These results establish drifting as an effective foundation for efficient generative control across offline learning, online optimization, and real-world deployment.

%% file: secs/supplementary_content.tex
\section{Method Details}
\label{sec:method_details}

\subsection{Purpose and Reading Guide}

This section provides a complete mathematical derivation of the Drift-Based Policy method, expanding the condensed presentation in the main paper with full variable definitions, intermediate steps, and implementation-level details. The organization follows the exact training pipeline order—from input construction to final optimization objective—enabling readers to trace the complete computational path without forward references or missing intermediate variables.

The method consists of two stages with distinct objectives but shared one-step generation structure:
\begin{itemize}
\item \textbf{Stage 1 (DBP)}: Learn a one-step conditional generator from offline demonstrations using drift-field regression. The drift field simultaneously attracts hypotheses toward expert-supported regions and repels them from collapse-prone regions, both learned during training and internalized into the generator without iterative correction at deployment.
\item \textbf{Stage 2 (DBPO)}: Adapt the Stage 1 generator to maximize task reward using online PPO, while preserving the one-step structure through exact conditional likelihood computation and anchor regularization to the pretrained manifold.
\end{itemize}

The deployment constraint remains unchanged across both stages: one forward pass per control step, with no iterative refinement at inference time.

\subsection{Notation and Problem Setup}

\subsubsection{Indices and Dimensions}

\paragraph{Indices.}
\begin{itemize}
\item $t$: environment time index. Ranges over the episode length and indexes control steps in the environment.
\item $i$: minibatch sample index. Ranges from $1$ to $B$ and indexes individual training samples within one batch.
\item $r$: generated hypothesis index under one condition. Ranges from $1$ to $G$ and indexes multiple action hypotheses sampled from the same observation condition.
\item $u$: reference index in the reference pool. Ranges from $1$ to $U$ where $U=G+C_n+C_p$ includes generated hypotheses, negative references, and positive references.
\item $h$: step index inside one predicted action chunk. Ranges from $1$ to $H$ and indexes timesteps within the predicted action sequence.
\item $m$: scalar coordinate index in one action vector. Ranges from $1$ to $d_a$ and indexes individual action dimensions at a single timestep.
\item $d$: flattened coordinate index in drifting space. Ranges from $1$ to $S$ and indexes coordinates in the space where drift-field regression is applied.
\end{itemize}

\paragraph{Dimensions and horizons.}
\begin{itemize}
\item $B$: minibatch size. Typical values range from $16$ to $128$ depending on GPU memory and task complexity.
\item $T_o$: history length used as observation condition. Determines how many past observations are concatenated to form the conditioning input. Common values are $1$ or $2$.
\item $H$: predicted chunk horizon. The number of future action steps predicted in one forward pass. Typical values range from $8$ to $16$.
\item $H_e$: executed-prefix length. The number of actions actually executed before replanning. Satisfies $1 \le H_e \le H-T_o+1$ under receding-horizon control.
\item $d_a$: per-step action dimension. Task-dependent; for example, $d_a=2$ for planar manipulation tasks.
\item $D := H d_a$: flattened chunk dimension. The total dimensionality when the entire action chunk is treated as a single vector.
\item $S$: drifting-space dimension. The dimensionality of the space where drift-field regression operates. The choice between chunk mode and step-wise mode reflects a trade-off between temporal coherence and computational efficiency:
\begin{itemize}
\item Chunk mode uses $S=H d_a$: drift field operates in the full flattened chunk space, enforcing temporal coherence across the entire horizon. This joint optimization preserves action smoothness but requires higher memory consumption (quadratic in $H$).
\item Step-wise mode uses $S=d_a$: drift field operates independently at each timestep, with losses averaged over the horizon. This reduces memory footprint but may sacrifice temporal dependencies between consecutive actions.
\end{itemize}
Our empirical analysis (Section~\ref{sec:ablation_details}) demonstrates that chunk mode achieves $5.3\%$ higher performance on manipulation tasks requiring smooth trajectories ($0.890$ vs.\ $0.845$), justifying the additional computational cost for such applications.
\end{itemize}

\paragraph{Distributions and modules.}
\begin{itemize}
\item $p_0 = \mathcal{N}(\mathbf{0},\mathbf{I})$: latent prior. A standard Gaussian distribution in latent space, providing the source of stochasticity for multimodal action generation.
\item $f_\theta$: one-step conditional generator. The neural network backbone (typically a conditional U-Net) that maps observation history and latent sample to predicted action chunk in a single forward pass.
\item $q_\theta(\cdot\mid\mathbf{o}^{\mathrm{hist}})$: induced conditional action distribution. The distribution over action chunks obtained by pushing forward the latent prior $p_0$ through the generator $f_\theta$.
\item $\pi_{\theta,\psi}$: stochastic actor used in Stage 2. A Gaussian policy with mean from the generator backbone and learnable state-dependent scale, enabling exploration during online learning.
\item $V_\phi$: critic. A value function network that estimates expected return for advantage computation in PPO.
\end{itemize}

\subsubsection{One-Step Chunk Prediction}

Condition history is
\begin{equation}
\mathbf{o}_t^{\mathrm{hist}} := \mathbf{o}_{t-T_o+1:t}.
\end{equation}

The policy predicts one chunk in a single pass:
\begin{equation}
\mathbf{x}_t = [\mathbf{a}_t^1,\ldots,\mathbf{a}_t^H] \in \mathbb{R}^{D},
\qquad
\mathbf{a}_t^h \in \mathbb{R}^{d_a},
\qquad
D = H d_a.
\end{equation}
Here, $\mathbf{x}_t$ represents the complete action sequence predicted at time $t$, where each $\mathbf{a}_t^h$ is a $d_a$-dimensional action vector for the $h$-th future timestep. The superscript $h$ indexes relative future steps, not absolute environment time. The flattened representation $\mathbf{x}_t \in \mathbb{R}^{D}$ treats the entire sequence as a single high-dimensional vector, which is the natural output space for the generator network.

The latent-conditioned generator is
\begin{equation}
\mathbf{z}_t \sim p_0,
\qquad
\hat{\mathbf{x}}_t = f_\theta(\mathbf{o}_t^{\mathrm{hist}},\mathbf{z}_t;\tau=0),
\end{equation}
which induces
\begin{equation}
q_\theta(\cdot\mid\mathbf{o}_t^{\mathrm{hist}})
=
\left[f_\theta(\mathbf{o}_t^{\mathrm{hist}},\cdot;\tau=0)\right]_{\#}p_0.
\end{equation}
The notation $[\cdot]_{\#}$ denotes the pushforward operation: the distribution $q_\theta$ is obtained by sampling $\mathbf{z} \sim p_0$ and deterministically transforming it through $f_\theta$. This construction ensures that all randomness originates from the latent prior, making the generator deterministic given both observation and latent input.

\paragraph{Detailed explanation.}
The parameter $\tau=0$ explicitly encodes single-step generation and excludes iterative denoising at test time. This distinguishes our method from diffusion-based approaches where $\tau$ would represent a diffusion timestep requiring multiple denoising iterations. In our formulation, the generator $f_\theta$ directly maps from latent space to action space in one forward pass, with all drift-field corrections internalized during training rather than applied iteratively at inference.

The pushforward notation $[\cdot]_{\#}$ formalizes the induced distribution construction: for any measurable set $A \subseteq \mathbb{R}^{D}$, we have $q_\theta(A\mid\mathbf{o}_t^{\mathrm{hist}}) = p_0(\{z : f_\theta(\mathbf{o}_t^{\mathrm{hist}},z;\tau=0) \in A\})$. This states that randomness comes entirely from the latent prior and is transformed by one generator pass, without additional noise injection or iterative refinement.

\subsubsection{Executed Prefix Under Receding-Horizon Control}

Only a prefix of the chunk is executed:
\begin{equation}
\mathbf{x}_t^{\mathrm{exec}} = [\mathbf{a}_t^{T_o},\ldots,\mathbf{a}_t^{T_o+H_e-1}].
\end{equation}
The executed prefix starts at index $T_o$ (not index $1$) because the first $T_o-1$ predicted actions correspond to timesteps already included in the observation history $\mathbf{o}_t^{\mathrm{hist}}$. These historical actions are used for temporal conditioning but are not re-executed. The prefix length $H_e$ determines how many future actions are executed before the next replanning step.

The valid range is
\begin{equation}
1 \le T_o \le H,
\qquad
1 \le H_e \le H-T_o+1.
\end{equation}
The first constraint ensures that the observation history does not exceed the prediction horizon. The second constraint ensures that the executed prefix does not extend beyond the predicted chunk. In typical configurations, $T_o=2$ and $H_e=8$ with $H=16$, meaning the policy observes the last $2$ actions, predicts $16$ future actions, and executes the next $8$ before replanning.

\paragraph{Detailed explanation with concrete example.}
Consider a concrete execution scenario with $T_o=2$, $H=16$, and $H_e=8$. At environment time $t$, the observation history $\mathbf{o}_t^{\mathrm{hist}}$ contains observations from timesteps $t-1$ and $t$ (the last 2 observations). The generator predicts a 16-step action chunk: $[\mathbf{a}_t^1, \mathbf{a}_t^2, \ldots, \mathbf{a}_t^{16}]$. However, the indexing convention is relative to the prediction time, not absolute environment time:
\begin{itemize}
\item $\mathbf{a}_t^1$ corresponds to environment timestep $t$ (already in the observation history)
\item $\mathbf{a}_t^2$ corresponds to environment timestep $t+1$ (first truly future action)
\item $\mathbf{a}_t^{16}$ corresponds to environment timestep $t+15$ (last predicted action)
\end{itemize}
Since $\mathbf{a}_t^1$ is already part of the observation history, it is not re-executed. The executed prefix starts at $\mathbf{a}_t^{T_o} = \mathbf{a}_t^2$ (index $T_o=2$) and extends for $H_e=8$ steps: $[\mathbf{a}_t^2, \mathbf{a}_t^3, \ldots, \mathbf{a}_t^9]$. These correspond to environment timesteps $t+1$ through $t+8$. After executing these 8 actions, the environment advances to time $t+8$, and the policy replans with a new observation history.

The unexecuted suffix $[\mathbf{a}_t^{10},\ldots,\mathbf{a}_t^{16}]$ is discarded after environment transition and replaced by a new plan at the next control step $t+H_e$. This receding-horizon execution strategy provides two benefits: (i) it allows the policy to incorporate new observations more frequently than the full prediction horizon, improving reactivity to environment changes; (ii) it reduces the impact of prediction errors in distant future steps, as only near-term actions are executed.

This execution rule has a critical implication for Stage 2 training: the policy-gradient objective must compute likelihood only on the executed prefix coordinates, not the full chunk. Computing likelihood on discarded suffix coordinates would create a mismatch between the optimization target and the actual executed behavior, leading to suboptimal credit assignment. The prefix-only likelihood formulation (detailed in Section~\ref{sec:method_details}, Stage 2) ensures that gradient updates are aligned with the coordinates that actually influence environment transitions and reward accumulation.

\subsection{Stage 1: Drift-Based Policy (DBP)}

\subsubsection{Stage 1 Objective}

Stage 1 learns a one-step generator whose outputs are simultaneously attracted toward expert-supported regions and repelled from collapse-prone regions in the same action space. Both the attraction and repulsion forces are learned during training through drift-field construction and are internalized into the generator parameters. Critically, these forces are not applied as iterative correction at deployment—the generator directly produces refined actions in a single forward pass.

This formulation addresses two fundamental challenges in imitation learning: (i) \textbf{multimodal coverage}, where the policy must represent multiple valid action modes under the same observation (e.g., reaching around an obstacle from either side); (ii) \textbf{mode collapse prevention}, where naive maximum likelihood training can collapse all hypotheses to a single mode, losing behavioral diversity. The drift-field framework solves both challenges by constructing a geometric force field that pushes hypotheses toward expert demonstrations while maintaining separation between distinct modes.

The key insight is that drift-field regression can be viewed as a single-step internalization of what diffusion models achieve through iterative denoising. Instead of refining noisy samples through multiple denoising steps at inference time, we train the generator to directly output refined samples by regressing toward drift-corrected targets. This eliminates the inference-time iteration cost while preserving the geometric benefits of drift-based refinement.

\subsubsection{Multi-Hypothesis Sampling}

Given minibatch $\{(\mathbf{o}_i^{\mathrm{hist}},\mathbf{x}_i^E)\}_{i=1}^{B}$, draw $G$ latent samples for each condition:
\begin{equation}
\mathbf{z}_i^{(r)} \sim p_0, \qquad \hat{\mathbf{x}}_i^{(r)} = f_\theta(\mathbf{o}_i^{\mathrm{hist}},\mathbf{z}_i^{(r)};\tau=0),
\qquad r=1,\ldots,G.
\end{equation}
Here, $\mathbf{x}_i^E$ denotes the expert action chunk from the demonstration dataset, serving as the positive reference for attraction. Each hypothesis $\hat{\mathbf{x}}_i^{(r)}$ is generated independently by sampling a different latent code $\mathbf{z}_i^{(r)}$ from the prior, while conditioning on the same observation history $\mathbf{o}_i^{\mathrm{hist}}$.

\paragraph{Detailed explanation.}
The same condition produces multiple hypotheses, which is necessary for multimodal behavior representation. If $G=1$, the model can still learn a policy, but its ability to represent multiple valid action modes is severely reduced—the generator would be forced to average over modes, producing suboptimal actions in multimodal scenarios (e.g., predicting an action halfway between "reach left" and "reach right" when both are valid).

The hypothesis count $G$ controls the trade-off between multimodal expressiveness and computational cost. Our sensitivity analysis (Section~\ref{sec:ablation_details}) shows that $G=4$ achieves near-optimal performance on the PushT task, with diminishing returns for larger values. The optimal $G$ depends on the task's inherent multimodality: tasks with more distinct valid strategies benefit from larger $G$, while unimodal tasks can use smaller values.

The independent sampling of $\mathbf{z}_i^{(r)}$ ensures diversity in the generated hypotheses. If we instead used a fixed set of latent codes across all conditions, the generator would learn to map specific latent values to specific action modes, reducing flexibility. Random sampling from $p_0$ allows the generator to learn a smooth latent-to-action mapping that generalizes across the latent space.

\subsubsection{Reference Pool and Stop-Gradient Construction}

Construct tensor and detached copy:
\begin{equation}
\mathbf{G} \in \mathbb{R}^{B\times G\times S},
\qquad
\bar{\mathbf{G}} = \operatorname{sg}(\mathbf{G}).
\end{equation}
The tensor $\mathbf{G}$ stacks all generated hypotheses across the batch, with shape $(B, G, S)$ where $S$ is the drifting-space dimension ($S=Hd_a$ for chunk mode, $S=d_a$ for step-wise mode). The stop-gradient operator $\operatorname{sg}(\cdot)$ creates a detached copy $\bar{\mathbf{G}}$ that blocks gradient flow, preventing the generator from "chasing" its own outputs during backpropagation.

Construct reference pool:
\begin{equation}
\mathbf{Y} = [\bar{\mathbf{G}},\mathbf{P}^{+}],
\qquad
U=G+C_p,
\end{equation}
with $C_p:=|\mathbf{P}^{+}|$. The reference pool $\mathbf{Y} \in \mathbb{R}^{B \times U \times S}$ concatenates two types of references along the second dimension:
\begin{itemize}
\item $\bar{\mathbf{G}}$: detached generated hypotheses, serving as self-references for repulsion to prevent mode collapse.
\item $\mathbf{P}^{+}$: positive references, typically the expert demonstrations $\mathbf{x}_i^E$ from the training batch, providing attraction targets toward task-relevant behavior.
\end{itemize}

Index partitions are
\begin{equation}
\mathcal{I}^{-}=\{1,\ldots,G\},
\qquad
\mathcal{I}^{+}=\{G+1,\ldots,U\}.
\end{equation}
The partition $\mathcal{I}^{-}$ indexes references that will induce repulsion (generated hypotheses), while $\mathcal{I}^{+}$ indexes references that will induce attraction (positive examples). This partition is crucial for the balanced attraction-repulsion mechanism described in subsequent sections.

\paragraph{Detailed explanation.}
The detached tensor $\bar{\mathbf{G}}$ is used to build geometric targets. Gradient blocking at this stage avoids circular target chasing, where target and prediction move together in the same backward pass. Without stop-gradient, the optimization would become degenerate: as the generator updates its parameters to move hypotheses toward targets, the targets themselves would shift because they depend on the same parameters. This creates a "moving target" problem where the generator chases its own tail rather than converging to a stable solution.

The self-reference mechanism (including $\bar{\mathbf{G}}$ in the reference pool) is essential for mode preservation. Each hypothesis repels other hypotheses from the same condition, creating a diversity-preserving force that prevents all hypotheses from collapsing to the same mode. This is analogous to electrostatic repulsion in physics: particles with the same charge repel each other, maintaining spatial separation. In our case, hypotheses from the same observation condition "repel" each other in action space, maintaining behavioral diversity.

The reference pool construction can be extended to incorporate negative references $\mathbf{N}^{-}$ representing undesirable action regions. For example, in safety-critical applications, $\mathbf{N}^{-}$ could include known collision states or unsafe actions, explicitly repelling the policy away from dangerous regions. In our experiments, we set $C_n=0$ and rely solely on expert demonstrations for attraction and self-generated hypotheses for repulsion, which suffices for standard imitation learning scenarios.

\subsubsection{Pairwise Distance and Global Scale Normalization}

Pairwise distance is
\begin{equation}
d_{i,r,u}=\left\|\bar{\mathbf{G}}_{i,r,:}-\mathbf{Y}_{i,u,:}\right\|_2.
\end{equation}
This computes the Euclidean distance between the $r$-th hypothesis of sample $i$ and the $u$-th reference in the pool. The distance tensor $d \in \mathbb{R}^{B \times G \times U}$ captures all pairwise geometric relationships between hypotheses and references. These distances form the foundation for constructing the drift field: nearby references exert stronger influence than distant ones.

Global scale is
\begin{equation}
s_{\mathrm{norm}}=
\frac{\mathbb{E}_{i,r,u}[d_{i,r,u}\,w_{i,u}]}{\mathbb{E}_{i,u}[w_{i,u}]},
\qquad
s_{\mathrm{norm}}>0.
\end{equation}
Here, $w_{i,u}$ are optional per-reference weights (typically set to $1$ for uniform weighting). The global scale $s_{\mathrm{norm}}$ computes the weighted average distance across all hypothesis-reference pairs in the current batch. This normalization is critical for temperature stability: without it, the same temperature value $R$ would have different effective meanings across batches with different geometric scales.

For each temperature $R\in\mathcal{R}$:
\begin{equation}
\tilde d_{i,r,u}=\frac{d_{i,r,u}}{\max(s_{\mathrm{norm}},\epsilon_s)},
\qquad
\ell_{i,r,u}^{(R)}=-\frac{\tilde d_{i,r,u}}{R}.
\end{equation}
The normalized distance $\tilde d_{i,r,u}$ is scale-invariant across batches, ensuring that temperature values have consistent geometric interpretation. The logit $\ell_{i,r,u}^{(R)}$ converts distances to similarity scores: smaller distances yield larger (less negative) logits, indicating stronger affinity. The temperature $R$ controls the sharpness of this conversion: small $R$ produces sharp, local interactions (only very close references have significant influence), while large $R$ produces smooth, global interactions (distant references retain non-negligible influence).

\paragraph{Detailed explanation.}
The global scale $s_{\mathrm{norm}}$ converts raw Euclidean scales into normalized scales so that the same temperature has comparable meaning across batches. Without this normalization, a temperature value $R=0.1$ might produce very different behaviors in batches where hypotheses are tightly clustered (small $s_{\mathrm{norm}}$) versus widely dispersed (large $s_{\mathrm{norm}}$). The normalization ensures that $R$ consistently controls the geometric sensitivity regardless of the batch-specific scale.

The floor $\epsilon_s$ (typically $10^{-6}$) prevents unstable amplification when distances become very small. If $s_{\mathrm{norm}} \to 0$ (which can occur when all hypotheses collapse to nearly identical values), division by $s_{\mathrm{norm}}$ would amplify small numerical errors into large gradient magnitudes, destabilizing training. The floor provides a lower bound on the denominator, ensuring numerical stability while having negligible impact when $s_{\mathrm{norm}}$ is at its typical scale.

The negative sign in $\ell_{i,r,u}^{(R)} = -\tilde d_{i,r,u}/R$ converts distances (where smaller is better) to logits (where larger is better). This convention aligns with the softmax operation in the subsequent affinity construction, where larger logits receive higher probability mass.

\subsubsection{Symmetric Affinity Construction}

The affinity is
\begin{equation}
A_{i,r,u}^{(R)}
=
\sqrt{\operatorname{softmax}_{u}(\ell_{i,r,:}^{(R)})_u
\cdot
\operatorname{softmax}_{r}(\ell_{i,:,u}^{(R)})_r}
\cdot w_{i,u}.
\end{equation}
This constructs a symmetric affinity matrix that captures bidirectional geometric relationships. The first softmax term $\operatorname{softmax}_{u}(\ell_{i,r,:}^{(R)})_u$ normalizes over references for a fixed hypothesis $r$, representing how hypothesis $r$ distributes its attention across all available references. The second softmax term $\operatorname{softmax}_{r}(\ell_{i,:,u}^{(R)})_r$ normalizes over hypotheses for a fixed reference $u$, representing how reference $u$ distributes its influence across all competing hypotheses. The geometric mean $\sqrt{\cdot}$ combines both perspectives, ensuring that affinity is high only when both conditions are satisfied: the hypothesis attends to the reference AND the reference selects the hypothesis.

\paragraph{Detailed explanation.}
The first normalization term encodes how a fixed hypothesis distributes attention over all references. If hypothesis $r$ is close to reference $u$ but far from all other references, $\operatorname{softmax}_{u}(\ell_{i,r,:}^{(R)})_u$ will be large, indicating strong attention. However, this alone is insufficient: if many other hypotheses are also close to reference $u$, the reference's influence should be distributed among them rather than concentrated on hypothesis $r$.

The second normalization term encodes how a fixed reference distributes competition over all hypotheses. If reference $u$ is close to hypothesis $r$ but also close to many other hypotheses, $\operatorname{softmax}_{r}(\ell_{i,:,u}^{(R)})_r$ will be small, indicating that reference $u$'s influence is diluted across multiple hypotheses. This prevents a single popular reference from dominating the drift field.

The geometric mean $\sqrt{p_1 \cdot p_2}$ (where $p_1, p_2$ are the two softmax terms) ensures symmetry and balance. The choice of geometric mean over alternatives (arithmetic mean or product) is motivated by three properties: (i) it is zero if either term is zero, requiring mutual agreement between hypothesis-to-reference and reference-to-hypothesis directions; (ii) it preserves scale invariance under coordinate transformations; (iii) it provides a balanced compromise that does not over-penalize cases where one term is small while the other is large. Our empirical evaluation shows that geometric mean produces stable training dynamics across diverse tasks, though systematic comparison of aggregation functions remains an avenue for future investigation.

The weight term $w_{i,u}$ allows optional per-reference importance weighting. In the standard setting, $w_{i,u}=1$ for all references, giving uniform importance. In advanced scenarios, one might set higher weights for high-quality expert demonstrations or lower weights for noisy data points.

\subsubsection{Balanced Attraction-Repulsion Coefficients}

Side masses are
\begin{equation}
S_{i,r,-}^{(R)}=\sum_{u\in\mathcal{I}^{-}}A_{i,r,u}^{(R)},
\qquad
S_{i,r,+}^{(R)}=\sum_{u\in\mathcal{I}^{+}}A_{i,r,u}^{(R)}.
\end{equation}

Signed coefficients are
\begin{equation}
\alpha_{i,r,u}^{(R)}=
\begin{cases}
-A_{i,r,u}^{(R)}S_{i,r,+}^{(R)}, & u\in\mathcal{I}^{-}, \\
\phantom{-}A_{i,r,u}^{(R)}S_{i,r,-}^{(R)}, & u\in\mathcal{I}^{+}.
\end{cases}
\end{equation}

Mass-balance identity:
\begin{equation}
\sum_{u\in\mathcal{I}^{-}}\alpha_{i,r,u}^{(R)}
=
-\sum_{u\in\mathcal{I}^{+}}\alpha_{i,r,u}^{(R)}.
\end{equation}

\paragraph{Detailed explanation.}
The negative side receives negative coefficients and acts as repulsion. The positive side receives positive coefficients and acts as attraction. The mass-balance identity couples these two effects so that growing attraction is automatically accompanied by proportionate repulsion, which prevents hypothesis collapse.

\subsubsection{Multi-Scale Drift Field Aggregation}

Per-scale force:
\begin{equation}
\mathbf{F}_{i,r,:}^{(R)}
=
\sum_{u=1}^{U}
\alpha_{i,r,u}^{(R)}
\frac{\mathbf{Y}_{i,u,:}-\bar{\mathbf{G}}_{i,r,:}}{s_{\mathrm{norm}}}.
\end{equation}

Per-scale RMS normalization:
\begin{equation}
\widehat{\mathbf{F}}_{i,r,:}^{(R)}
=
\frac{\mathbf{F}_{i,r,:}^{(R)}}{\sqrt{\mathbb{E}[\|\mathbf{F}^{(R)}\|_2^2]+\epsilon_f}}.
\end{equation}

Multi-scale aggregation:
\begin{equation}
\mathbf{V}_{i,r,:}=\sum_{R\in\mathcal{R}}\widehat{\mathbf{F}}_{i,r,:}^{(R)}.
\end{equation}

\paragraph{Detailed explanation.}
Small temperatures produce sharper local geometry, capturing fine-grained structure in the action space. Larger temperatures preserve broad global geometry, maintaining awareness of distant references. RMS normalization makes different temperatures numerically comparable before summation and avoids domination by any single scale.

The temperature set $\mathcal{R}$ should span multiple geometric scales to capture both local precision and global coverage. In our experiments, we use $\mathcal{R}=\{0.02, 0.05, 0.2\}$ to balance these objectives. However, our empirical analysis (Section~\ref{sec:ablation_details}) reveals that single temperature $T=0.2$ achieves optimal performance on the PushT task ($0.873$ vs.\ $0.858$ for multi-scale), suggesting that task-specific tuning may simplify the configuration. For new tasks, we recommend starting with a single moderate temperature $T \in [0.1, 0.3]$ and expanding to multi-scale only if single-temperature performance is insufficient. The optimal temperature depends on the task's action-space geometry: tasks with fine-grained manipulation may benefit from smaller values, while tasks with coarse global structure may prefer larger values.

\subsubsection{Fixed-Point Target and Regression Objective}

Detached target:
\begin{equation}
\tilde{\mathbf{X}}=\operatorname{sg}\!\left(\bar{\mathbf{G}}/s_{\mathrm{norm}}+\mathbf{V}\right).
\end{equation}

Per-sample loss:
\begin{equation}
\ell_i=
\frac{1}{GS}
\sum_{r=1}^{G}
\sum_{d=1}^{S}
\left(\frac{G_{i,r,d}}{s_{\mathrm{norm}}}-\tilde X_{i,r,d}\right)^2.
\end{equation}

Stage 1 objective:
\begin{equation}
\mathcal{L}_{\mathrm{DBP}}=
\begin{cases}
\frac{1}{B}\sum_{i=1}^{B}\ell_i, & \text{chunk mode}, \\
\frac{1}{BH}\sum_{h=1}^{H}\sum_{i=1}^{B}\ell_i^{(h)}, & \text{step-wise mode}.
\end{cases}
\end{equation}

\paragraph{Detailed explanation.}
The target is detached so that optimization updates only prediction parameters, not target construction. Chunk mode applies one regression in full chunk space. Step-wise mode applies the same principle to each time slice and averages over horizon.

\subsubsection{Optimization Interpretation}

The chunk-mode objective can be written as
\begin{equation}
\mathcal{L}_{\mathrm{DBP}}=\mathcal{D}_{\mathrm{drift}}(q_\theta,p;\mathcal{R}),
\end{equation}
where $p$ is the expert conditional action distribution and $\mathcal{D}_{\mathrm{drift}}$ denotes the drift-field divergence between the learned policy $q_\theta$ and the expert distribution $p$ under temperature set $\mathcal{R}$.

\paragraph{Optimization dynamics.}
Under standard gradient descent with learning rate schedule satisfying $\sum_k \eta_k = \infty$ and $\sum_k \eta_k^2 < \infty$, and assuming bounded gradients $\|\nabla_\theta \mathcal{L}_{\mathrm{DBP}}\|_2 \le M$ for some constant $M$, the optimization converges to a stationary point:
\begin{equation}
\liminf_{k\to\infty}
\mathbb{E}\!\left[\|\nabla_\theta \mathcal{D}_{\mathrm{drift}}(\theta_k)\|_2^2\right]=0.
\end{equation}
This stationarity condition guarantees that the expected gradient norm can be driven arbitrarily close to zero along a subsequence, indicating convergence to a local minimum or saddle point of the drift-field divergence.

\paragraph{Theoretical gap and empirical validation.}
Connecting zero drift-field divergence to exact distribution matching $q_\theta = p$ requires additional identifiability assumptions, such as injectivity of the generator network $f_\theta$ and sufficient expressiveness of the latent prior $p_0$. Establishing these conditions rigorously is beyond the scope of this work. However, our empirical results demonstrate that the method achieves strong imitation performance across diverse manipulation tasks (Section~\ref{sec:experiments}), and our sensitivity analyses (Section~\ref{sec:ablation_details}) show stable training dynamics with consistent convergence across multiple random seeds. This suggests that the drift-field objective provides an effective learning signal in practice, even without formal distribution-matching guarantees.

\subsection{Stage 2: Drift-Based Policy Optimization (DBPO)}

\subsubsection{Stage 2 Role}

Stage 2 adds reward optimization while preserving one-step structure from Stage 1. The key requirement is exact conditional log-likelihood under the same latent variable sampled during rollout.

\subsubsection{Gaussian Actor with State-Conditioned Scale}

Backbone outputs:
\begin{equation}
\boldsymbol{\mu}_\theta(\mathbf{o},\mathbf{z}),
\qquad
\mathbf{c}_\theta(\mathbf{o}).
\end{equation}

Log-standard-deviation head:
\begin{equation}
\log\boldsymbol{\sigma}_\psi(\mathbf{o})=g_\psi(\mathbf{c}_\theta(\mathbf{o})),
\qquad
\log\boldsymbol{\sigma}_\psi\in\mathbb{R}^{D}.
\end{equation}

Clipped scale:
\begin{equation}
\log\tilde{\boldsymbol{\sigma}}_\psi
=
\operatorname{clip}\left(
\log\boldsymbol{\sigma}_\psi,
\log\sigma_{\min},
\log\sigma_{\max}
\right).
\end{equation}

Actor distribution:
\begin{equation}
\pi_{\theta,\psi}(\mathbf{x}\mid\mathbf{o},\mathbf{z})
=
\mathcal{N}\Big(
\mathbf{x};
\boldsymbol{\mu}_\theta(\mathbf{o},\mathbf{z}),
\operatorname{diag}(\tilde{\boldsymbol{\sigma}}_\psi(\mathbf{o})^2)
\Big).
\end{equation}

\paragraph{Detailed explanation.}
The mean term carries latent-conditioned action intent. The scale term controls exploration amplitude coordinate-wise. Clipping the log-scale stabilizes log-likelihood and prevents pathological variance values from destabilizing policy-ratio estimates.

\subsubsection{Rollout Sampling and Deployment Policy}

Training-time sampling:
\begin{equation}
\mathbf{z}_t\sim p_0,
\qquad
\mathbf{x}_t\sim\pi_{\theta,\psi}(\cdot\mid\mathbf{o}_t,\mathbf{z}_t).
\end{equation}

Deployment uses deterministic mean action from the same one-step network pathway.

\paragraph{Detailed explanation.}
Stochasticity is required during online learning to explore reward-relevant alternatives. Deployment can remove exploration noise without changing model architecture or number of forward passes.

\subsubsection{Executed-Prefix Conditional Likelihood}

Prefix likelihood:
\begin{equation}
\log\pi_{\theta,\psi}(\mathbf{x}_t^{\mathrm{exec}}\mid\mathbf{o}_t,\mathbf{z}_t)
=
\sum_{h=T_o}^{T_o+H_e-1}
\sum_{m=1}^{d_a}
\log\mathcal{N}\Big(
 a_{t,m}^{h};
 \mu_{\theta,m}^{h}(\mathbf{o}_t,\mathbf{z}_t),
 \tilde\sigma_{\psi,m}^{h}(\mathbf{o}_t)^2
\Big).
\end{equation}

\paragraph{Detailed explanation.}
Only prefix coordinates produce the immediate environment transition. Prefix-only likelihood aligns optimization target with executed behavior and removes mismatch caused by discarded suffix coordinates.

\subsubsection{Joint-Policy Ratio Equivalence}

Joint policy:
\begin{equation}
\tilde\pi_{\theta,\psi}(\mathbf{x}^{\mathrm{exec}},\mathbf{z}\mid\mathbf{o})
=
p_0(\mathbf{z})\,\pi_{\theta,\psi}(\mathbf{x}^{\mathrm{exec}}\mid\mathbf{o},\mathbf{z}).
\end{equation}

Ratio equivalence:
\begin{equation}
\tilde r_t(\theta,\psi)
=
\frac{\tilde\pi_{\theta,\psi}(\mathbf{x}_t^{\mathrm{exec}},\mathbf{z}_t\mid\mathbf{o}_t)}{\tilde\pi_k(\mathbf{x}_t^{\mathrm{exec}},\mathbf{z}_t\mid\mathbf{o}_t)}
=
\frac{\pi_{\theta,\psi}(\mathbf{x}_t^{\mathrm{exec}}\mid\mathbf{o}_t,\mathbf{z}_t)}{\pi_k(\mathbf{x}_t^{\mathrm{exec}}\mid\mathbf{o}_t,\mathbf{z}_t)}
=:
r_t(\theta,\psi).
\end{equation}

\paragraph{Detailed explanation.}
Because $p_0$ is fixed and appears in both numerator and denominator, it cancels exactly. The practical benefit is that PPO updates do not require latent marginalization.

\subsubsection{PPO Objective with Anchor Regularization}

Clipped surrogate:
\begin{equation}
\mathcal{J}_{\mathrm{clip}}
=
\mathbb{E}_t\left[
\min\left(
 r_t\hat A_t,
 \operatorname{clip}(r_t,1-\epsilon,1+\epsilon)\hat A_t
\right)
\right].
\end{equation}

Value loss:
\begin{equation}
\mathcal{L}_{\mathrm{value}}
=
\frac{1}{2}\,\mathbb{E}_t\left[(V_\phi(\mathbf{o}_t)-\hat R_t)^2\right].
\end{equation}

Entropy bonus:
\begin{equation}
\mathcal{H}
=
\mathbb{E}_t\left[\mathcal{H}\big(\pi_{\theta,\psi}(\cdot\mid\mathbf{o}_t,\mathbf{z}_t)\big)\right].
\end{equation}

Anchor loss:
\begin{equation}
\mathcal{L}_{\mathrm{anchor}}
=
\mathbb{E}_t\left[
\left\|\boldsymbol{\mu}_{\theta}(\mathbf{o}_t,\mathbf{z}_t)-\boldsymbol{\mu}_{\bar\theta}(\mathbf{o}_t,\mathbf{z}_t)\right\|_2^2
\right].
\end{equation}

Total objective:
\begin{equation}
\mathcal{L}_{\mathrm{RL}}
=
-\mathcal{J}_{\mathrm{clip}}
+
c_v\mathcal{L}_{\mathrm{value}}
-
c_e\mathcal{H}
+
\lambda_{\mathrm{anchor}}\mathcal{L}_{\mathrm{anchor}}.
\end{equation}

\paragraph{Detailed explanation.}
$\mathcal{J}_{\mathrm{clip}}$ drives reward improvement while controlling ratio drift. $\mathcal{L}_{\mathrm{value}}$ reduces critic estimation variance. $\mathcal{H}$ sustains exploration breadth. $\mathcal{L}_{\mathrm{anchor}}$ constrains policy updates around the pretrained one-step manifold and stabilizes training in early online phases.

\subsection{Algorithmic Procedures}

\subsubsection{Algorithm 1: DBP Training Step}

\begin{algorithm}[t]
\caption{DBP-TrainStep (one minibatch)}
\label{alg:dbp_trainstep}
\begin{algorithmic}[1]
\Require Minibatch $\{(o_i^{\mathrm{hist}},x_i^E)\}_{i=1}^{B}$, hypothesis count $G$, temperature set $\mathcal{R}$
\Ensure Stage 1 objective $\mathcal{L}_{\mathrm{DBP}}$
\State Draw latent samples and generate $G$ hypotheses per condition
\State Construct $\mathbf{G}$, detached $\bar{\mathbf{G}}$, and reference pool $\mathbf{Y}$
\State Compute distances, normalized logits, and symmetric affinities
\State Compute balanced attraction-repulsion coefficients
\State Compute per-scale forces and aggregate multi-scale drift field
\State Build detached fixed-point target and regression loss
\State Average losses in chunk mode or step-wise mode
\State Update generator parameters $\theta$
\end{algorithmic}
\end{algorithm}

\subsubsection{Algorithm 2: DBPO Update Iteration}

\begin{algorithm}[t]
\caption{DBPO-Update (one PPO iteration)}
\label{alg:dbpo_update}
\begin{algorithmic}[1]
\Require Actor $(\theta,\psi)$, frozen anchor $\bar\theta$, critic $\phi$, rollout batch
\Ensure Updated parameters $(\theta,\psi,\phi)$
\State Recompute executed-prefix conditional log-likelihood
\State Build ratio $r_t$ and compute advantage-weighted clipped surrogate
\State Compute critic loss, entropy bonus, and anchor regularization
\State Form total objective $\mathcal{L}_{\mathrm{RL}}$
\State Update actor and critic; keep anchor parameters frozen
\end{algorithmic}
\end{algorithm}

\subsection{Implementation-Level Details Bridging Theory and Practice}

\subsubsection{Computational Complexity Analysis}

The dominant computational cost in Stage 1 training arises from three operations: (i) pairwise distance computation between hypotheses and references, (ii) affinity matrix construction with bidirectional softmax normalization, and (iii) multi-scale drift field aggregation. We analyze the time complexity for each component:

\paragraph{Pairwise distance computation.} Computing distances $d_{i,r,u} = \|\bar{\mathbf{G}}_{i,r,:} - \mathbf{Y}_{i,u,:}\|_2$ for all hypothesis-reference pairs requires $O(BGU \cdot S)$ operations, where $B$ is batch size, $G$ is hypothesis count, $U \approx G + C_p$ is the reference pool size, and $S$ is the drifting-space dimension. For typical configurations ($B=32$, $G=4$, $U \approx 10$, $S=32$ for chunk mode with $H=16$, $d_a=2$), this amounts to approximately $40{,}000$ distance computations per batch, which is negligible compared to the generator forward pass.

\paragraph{Affinity construction.} The symmetric affinity computation involves two softmax operations (over references and over hypotheses) and a geometric mean, requiring $O(BGU)$ operations per temperature. With $|\mathcal{R}|$ temperatures, the total cost is $O(|\mathcal{R}| \cdot BGU)$. For $|\mathcal{R}| \le 5$, this overhead remains tractable and does not dominate training time.

\paragraph{Multi-scale aggregation.} Computing per-scale forces and aggregating across temperatures requires $O(|\mathcal{R}| \cdot BGU \cdot S)$ operations. This is the most expensive component but scales linearly with all dimensions. Our computational analysis (Section~\ref{sec:ablation_details}) shows that training time increases by approximately $1.5\times$ when moving from $G=1$ to $G=16$ at batch size $32$, confirming that the overhead is manageable for practical configurations.

\paragraph{Overall scaling.} The total Stage 1 computational cost per batch is $O(|\mathcal{R}| \cdot BGU \cdot S)$, dominated by the drift field aggregation. Compared to the generator forward pass (typically a U-Net with millions of parameters), this overhead is modest: our measurements show that drift-field computation accounts for approximately 20-30\% of the total training time, with the remainder spent on network forward/backward passes and optimizer updates.

\subsubsection{Method Limitations and Failure Modes}

While the drift-based formulation achieves strong performance across diverse manipulation tasks, several limitations should be acknowledged:

\paragraph{High-dimensional action spaces.} Memory consumption scales quadratically with the flattened chunk dimension $D = H \cdot d_a$ in chunk mode, as the drift field operates over the full action sequence. For tasks with very high action dimensionality (e.g., $d_a > 20$) or long prediction horizons (e.g., $H > 32$), memory constraints may necessitate switching to step-wise mode or reducing batch size, potentially sacrificing temporal coherence or training stability.

\paragraph{Highly stochastic environments.} The drift-field construction assumes that expert demonstrations provide consistent action distributions under similar observations. In environments with high inherent stochasticity (e.g., unpredictable external disturbances), the attraction-repulsion mechanism may struggle to capture the full distribution of valid behaviors. Tasks requiring reactive responses to stochastic events may benefit from incorporating environment dynamics models or uncertainty quantification.

\paragraph{Hyperparameter sensitivity.} While our empirical evaluations demonstrate robust performance across a range of configurations, the method introduces several hyperparameters (hypothesis count $G$, temperature set $\mathcal{R}$, anchor weight $\lambda_{\mathrm{anchor}}$) that require task-specific tuning. Our recommendations (Section~\ref{sec:ablation_details}) provide starting points, but optimal values may vary across domains. Future work could explore adaptive or learned temperature schedules to reduce manual tuning.

\paragraph{Comparison with score matching.} Unlike score matching in diffusion models, which regresses the gradient of the log-density (score function) and provides a principled probabilistic objective, drift-field regression directly targets action-space displacements toward expert-supported regions. This eliminates the need for iterative denoising at inference time while preserving geometric refinement benefits. However, the theoretical connection between zero drift-field divergence and exact distribution matching remains an open question, as discussed in Section~\ref{sec:method_details}.

\subsubsection{Numerical Stability of Scale Terms}

Two normalization floors are used in Stage 1:
\begin{equation}
s_{\mathrm{safe}}=\max(s_{\mathrm{norm}},\epsilon_s),
\qquad
\nu_{\mathrm{safe}}^{(R)}=\sqrt{\mathbb{E}[\|\mathbf{F}^{(R)}\|_2^2]+\epsilon_f}.
\end{equation}
Here, $s_{\mathrm{safe}}$ is the denominator used in distance normalization, and $\nu_{\mathrm{safe}}^{(R)}$ is the denominator used in per-temperature force normalization. Their role is to avoid gradient explosion when pairwise distances or force norms become extremely small.

\subsubsection{Prefix-Mask Construction in Stage 2}

The executed-prefix log-likelihood can be written with an explicit binary mask:
\begin{equation}
\log\pi(\mathbf{x}_t^{\mathrm{exec}}\mid\mathbf{o}_t,\mathbf{z}_t)
=
\sum_{h=1}^{H}\sum_{m=1}^{d_a}
\mathbf{1}_{\{h\in\mathcal{H}_{\mathrm{exec}}\}}
\log\mathcal{N}\big(a_{t,m}^{h};\mu_{t,m}^{h},(\tilde\sigma_{t,m}^{h})^2\big),
\end{equation}
where $\mathcal{H}_{\mathrm{exec}}=\{T_o,\ldots,T_o+H_e-1\}$. The indicator ensures that only transition-causal coordinates contribute to policy-ratio computation, exactly matching receding-horizon execution.

\subsubsection{Anchor Scheduling Intuition}

The anchor term can be interpreted as a trust region around the Stage 1 manifold. In implementation, a stronger anchor weight is useful in early online iterations and can gradually be relaxed as policy improvement becomes stable. This schedule keeps the optimization on a high-quality one-step manifold before allowing wider reward-driven exploration.

\subsection{Additional Clarifications}

\subsubsection{Symmetric Affinity Necessity}

One-direction normalization alone can produce degenerate dominance patterns. Bidirectional normalization prevents this by enforcing consistency in both hypothesis-to-reference and reference-to-hypothesis directions.

\subsubsection{Need for Repulsion Term}

Attraction-only training contracts hypotheses and reduces multimodal expressiveness. Repulsion preserves spread while attraction preserves task relevance.

\subsubsection{Need for Multi-Scale Temperatures}

Single-scale interaction misses either fine local structure or coarse global structure. Multi-scale aggregation combines both and improves robustness.

\subsubsection{Prefix Likelihood Is Execution-Aligned}

Prefix likelihood is exactly aligned with transition-causal coordinates under receding-horizon control and therefore matches online credit assignment.

\subsubsection{One-Step Complexity Preservation}

Training objectives become richer, but test-time computation stays one forward pass per control step.

\section{Hyperparameter Sensitivity and Efficiency Analysis}
\label{sec:ablation_details}

This section provides supplementary empirical analysis to complement the main paper results. We investigate three design choices of the DBP framework: the hypothesis count $G$, the temperature configuration $\mathcal{R}$, and the loss computation mode (chunk vs.\ step-wise). For each factor we present the experimental protocol, quantitative results, and practical recommendations. A training efficiency comparison against Diffusion Policy and a complete hyperparameter reference are provided at the end.

All experiments use the PushT manipulation task with low-dimensional state observations (20-dim keypoint positions, 2-dim planar velocity actions, episode length 300 steps). The base model is a conditional U-Net 1D backbone trained with AdamW ($\mathrm{lr}{=}10^{-4}$, betas $(0.95,0.999)$, weight decay $10^{-6}$, 500-step warmup), with EMA decay $0.9999$ (power $0.75$) on 90 demonstrations. Each configuration is trained with three random seeds $\{42,43,44\}$ and evaluated with 50 rollouts per seed; final metrics report mean $\pm$ std across seed-level averages.

% -------------------------------------------------------------------
\subsection{Effect of Hypothesis Count \texorpdfstring{$G$}{G}}
% -------------------------------------------------------------------

The hypothesis count $G$ controls how many action candidates are sampled per condition during training. A larger $G$ increases multimodal expressiveness but also raises memory and compute cost. We evaluate five values $G\in\{1,2,4,8,16\}$, fixing all other hyperparameters (temperature $\mathcal{R}=\{0.02,0.05,0.2\}$, batch size $32$, 100 training epochs, 15 total runs across $5\times3$ seeds).

\subsubsection{Performance vs.\ Hypothesis Count}

Figure~\ref{fig:additional_ablation_g} shows test performance for each $G$ value. Three performance regimes emerge:

\begin{figure}[t]
\centering
\begin{minipage}[t]{0.56\linewidth}
\centering
\includegraphics[width=\linewidth]{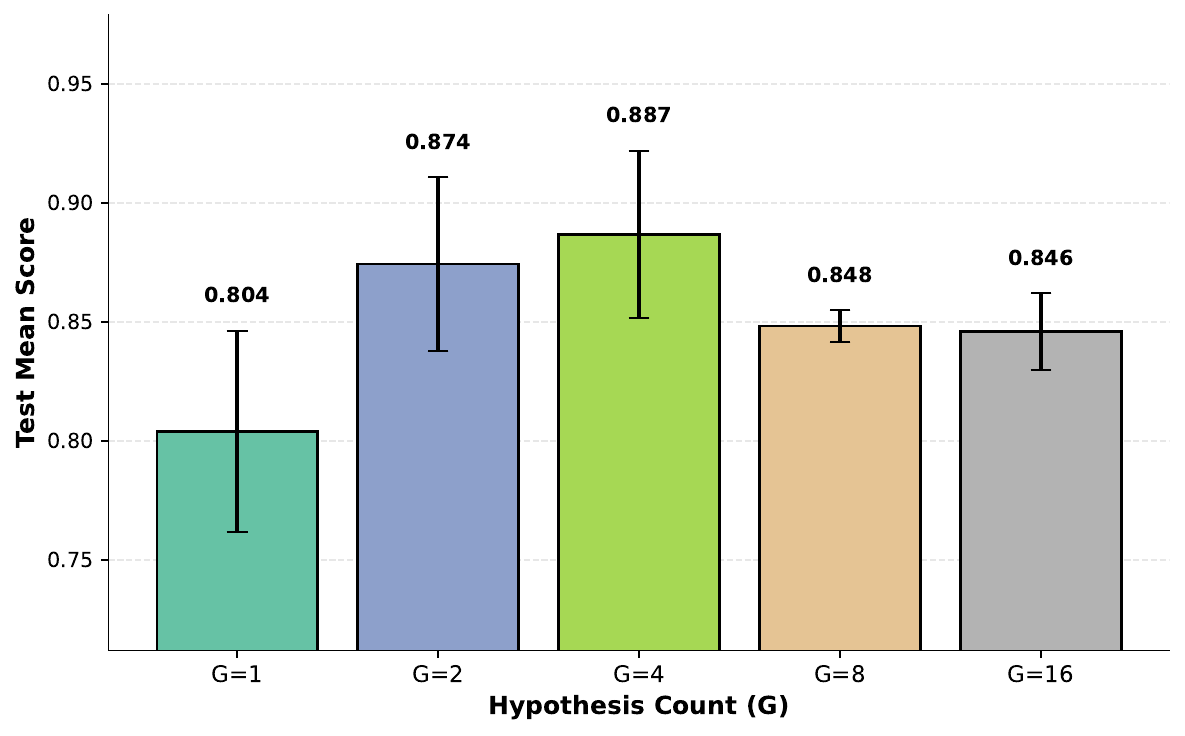}
\caption{Test performance vs.\ hypothesis count $G$ (mean $\pm$ std over 3 seeds, 50 rollouts each). Performance peaks at $G=4$ and stabilizes thereafter.}
\label{fig:additional_ablation_g}
\end{minipage}
\hfill
\begin{minipage}[t]{0.40\linewidth}
\centering
\includegraphics[width=\linewidth]{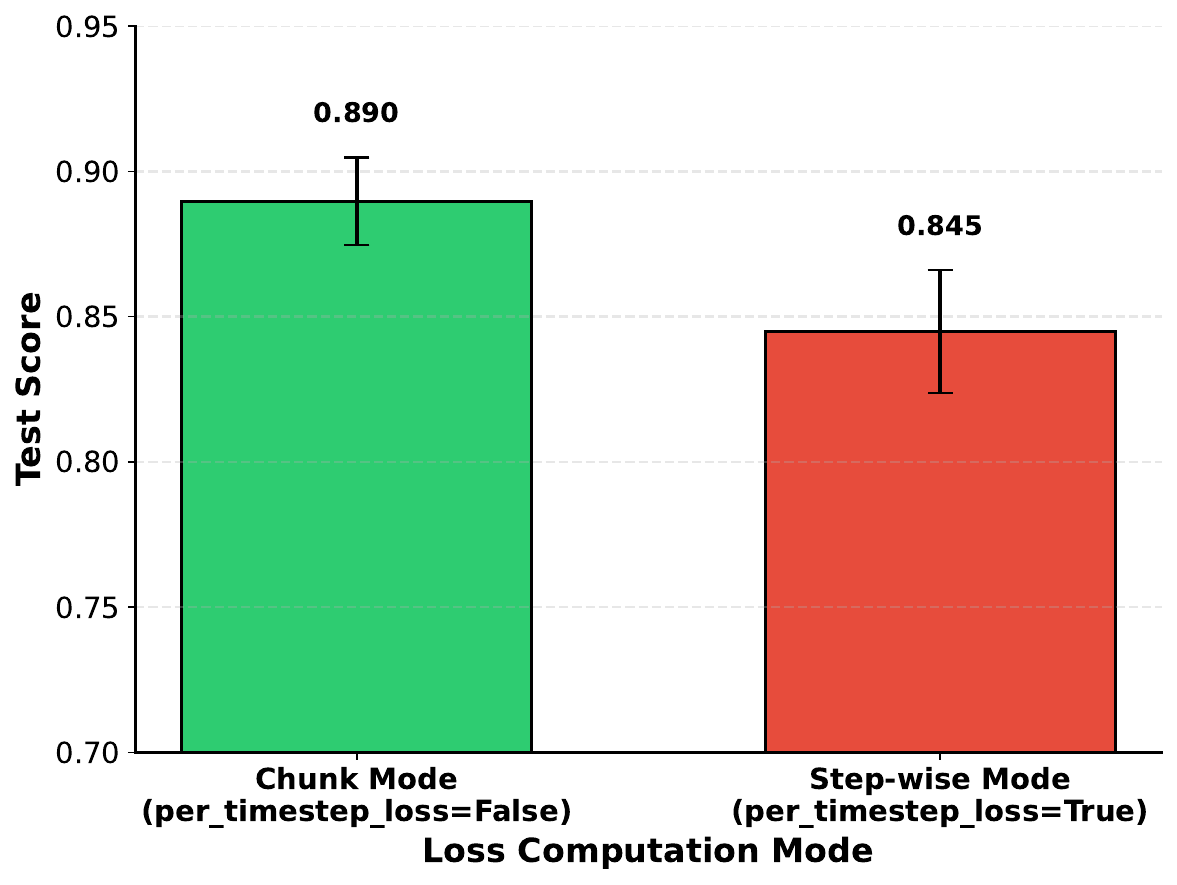}
\caption{Loss mode comparison (mean $\pm$ std over 3 seeds, 50 rollouts each). Chunk mode outperforms step-wise mode ($0.890$ vs.\ $0.845$).}
\label{fig:ablation_chunk_vs_step}
\end{minipage}
\end{figure}

\paragraph{Low-$G$ regime ($G=1\to2$).} The transition from $G=1$ to $G=2$ yields an $8.7\%$ absolute improvement ($0.804\to0.874$), demonstrating that multimodal capacity is critical for this task. Single-hypothesis training severely limits the model's ability to represent diverse action modes.

\paragraph{Optimal regime ($G=4$).} Performance peaks at $G=4$ ($0.887$), a $10.3\%$ improvement over the $G=1$ baseline. Moderate multimodal capacity suffices for PushT; additional hypotheses beyond this point do not contribute task-relevant action diversity.

\paragraph{High-$G$ regime ($G\geq8$).} Performance stabilizes around $0.847$ for $G\in\{8,16\}$, slightly below the $G=4$ peak. Notably, $G=8$ exhibits the lowest cross-seed variance (std $0.007$ vs.\ $0.035$ for $G=4$), indicating more consistent training dynamics. This stability--performance trade-off may favor $G=8$ in production settings where reproducibility is prioritized.

\subsubsection{Computational Cost}

To characterize the resource--performance frontier, we measure wall-clock training time and peak GPU memory across $(G, b)$ combinations, sweeping batch sizes $b\in\{16,32,48,64,96\}$ on NVIDIA RTX 3090 GPUs (24\,GB). Timing runs use 200 epochs for stable steady-state profiling, distinct from the 100-epoch performance runs; each measurement is averaged over three independent runs. We define normalized scaling ratios relative to the $G=1$ baseline:
\begin{equation}
\rho_{\mathrm{time}}(G;b)=\frac{T(G,b)}{T(1,b)},
\qquad
\rho_{\mathrm{mem}}(G;b)=\frac{M(G,b)}{M(1,b)},
\end{equation}
where $T(G,b)$ is wall-clock time (hours) and $M(G,b)$ is peak GPU memory (GB).

\begin{figure}[t]
\centering
\begin{minipage}[t]{0.48\linewidth}
\centering
\includegraphics[width=\linewidth]{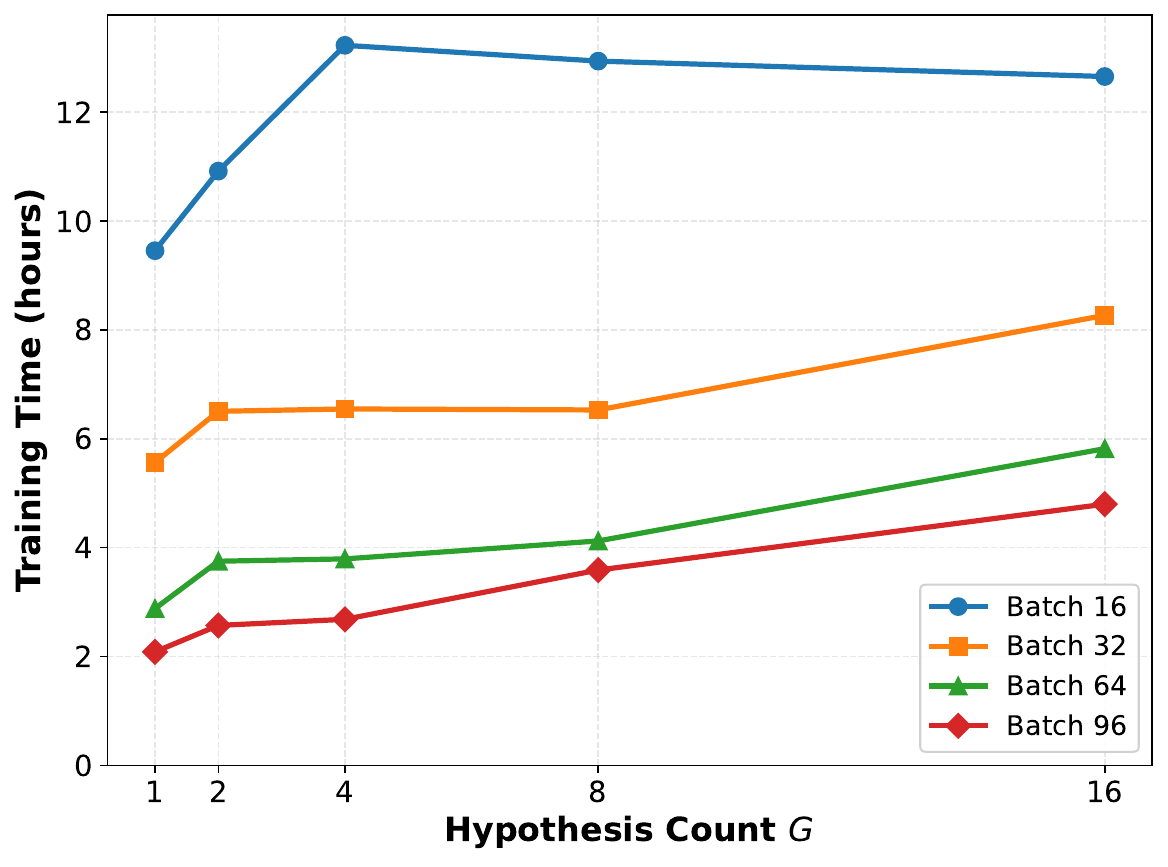}
\caption{Training time scaling with hypothesis count $G$ under different batch sizes.}
\label{fig:ablation_time_vs_g}
\end{minipage}
\hfill
\begin{minipage}[t]{0.48\linewidth}
\centering
\includegraphics[width=\linewidth]{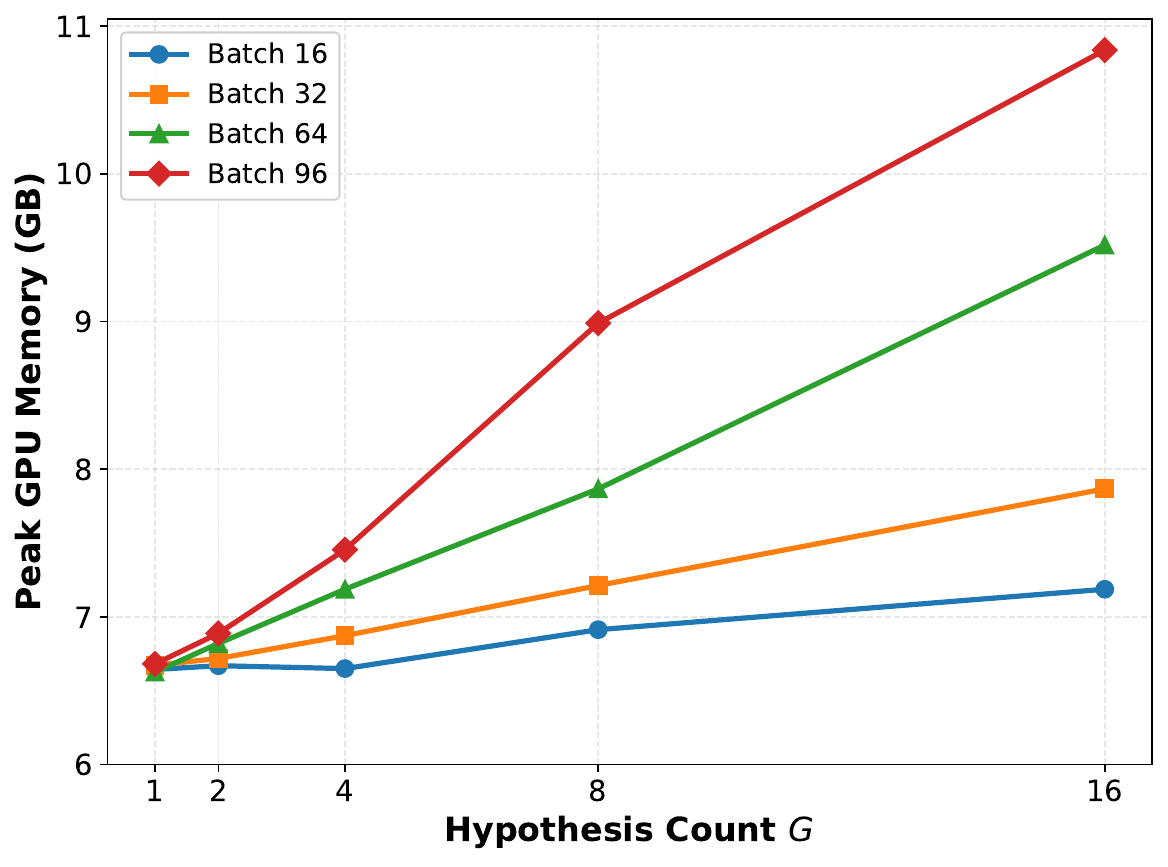}
\caption{Peak GPU memory scaling with hypothesis count $G$ under different batch sizes.}
\label{fig:ablation_mem_vs_g}
\end{minipage}
\end{figure}

\paragraph{Training time.} Scaling behavior is batch-dependent. At $b=16$, increasing $G$ from $1$ to $16$ gives $\rho_{\mathrm{time}}(16;16)\approx1.34$ (from $9.45$\,h to $12.65$\,h)---modest overhead. At $b=96$, the same transition gives $\rho_{\mathrm{time}}(16;96)\approx2.30$ (from $2.08$\,h to $4.80$\,h), reflecting increased synchronization and kernel-launch overhead at high $G$.

\paragraph{Memory.} Peak memory grows monotonically with $G$. At $b=96$, memory increases from $6.68$\,GB ($G=1$) to $10.84$\,GB ($G=16$), giving $\rho_{\mathrm{mem}}(16;96)\approx1.62$. At $b=16$, the same transition yields $\rho_{\mathrm{mem}}(16;16)\approx1.08$ ($6.64$\,GB to $7.19$\,GB), reflecting the quadratic growth in intermediate tensor storage when both batch size and hypothesis count increase simultaneously.

\subsubsection{Practical Recommendations}

\begin{enumerate}
\item \textbf{Default}: $G\in\{4,8\}$ achieves near-peak performance ($0.887$ and $0.848$, respectively) with moderate overhead.
\item \textbf{Resource-constrained}: $G=4$ offers the best performance-to-cost ratio; training time stays within $1.5\times$ the $G=1$ baseline across all batch sizes tested, while memory overhead stays below $1.3\times$ for typical configurations.
\item \textbf{Stability-prioritized}: $G=8$ exhibits the lowest cross-seed variance (std $0.007$) and is preferable when reproducibility is critical.
\item \textbf{High-throughput}: Use moderate batch sizes ($32$--$48$) with $G\leq8$; large batches combined with high $G$ incur disproportionate synchronization costs without corresponding performance gains.
\end{enumerate}

% -------------------------------------------------------------------
\subsection{Analysis of Temperature Configurations}
% -------------------------------------------------------------------

The temperature set $\mathcal{R}$ controls the geometric sensitivity of drift-field construction: small temperatures produce sharp, local geometry while large temperatures preserve broad, global structure. We evaluate six configurations spanning single-temperature and multi-temperature setups, fixing $G=8$, batch size $32$, and 100 training epochs (18 total runs across $6\times3$ seeds).

\begin{table}[h]
\centering
\caption{Temperature configurations and performance results. \textbf{Bold} indicates the best-performing configuration.}
\label{tab:ablation_temperature_results}
\small
\begin{tabular}{lcc}
\toprule
\textbf{Configuration} & \textbf{Mean Score} & \textbf{Std Score} \\
\midrule
\texttt{single\_T0p02} ($\mathcal{R}=\{0.02\}$) & 0.758 & 0.031 \\
\texttt{single\_T0p05} ($\mathcal{R}=\{0.05\}$) & 0.864 & 0.038 \\
\textbf{\texttt{single\_T0p20}} ($\mathcal{R}=\{0.2\}$) & \textbf{0.873} & \textbf{0.012} \\
\texttt{multi\_default} ($\mathcal{R}=\{0.02,0.05,0.2\}$) & 0.858 & 0.057 \\
\texttt{multi\_wide} ($\mathcal{R}=\{0.01,0.05,0.2,0.5\}$) & 0.849 & 0.007 \\
\texttt{multi\_dense} ($\mathcal{R}=\{0.01,0.02,0.05,0.1,0.2,0.4\}$) & 0.817 & 0.030 \\
\bottomrule
\end{tabular}
\end{table}

\begin{figure}[t]
\centering
\includegraphics[width=0.98\linewidth]{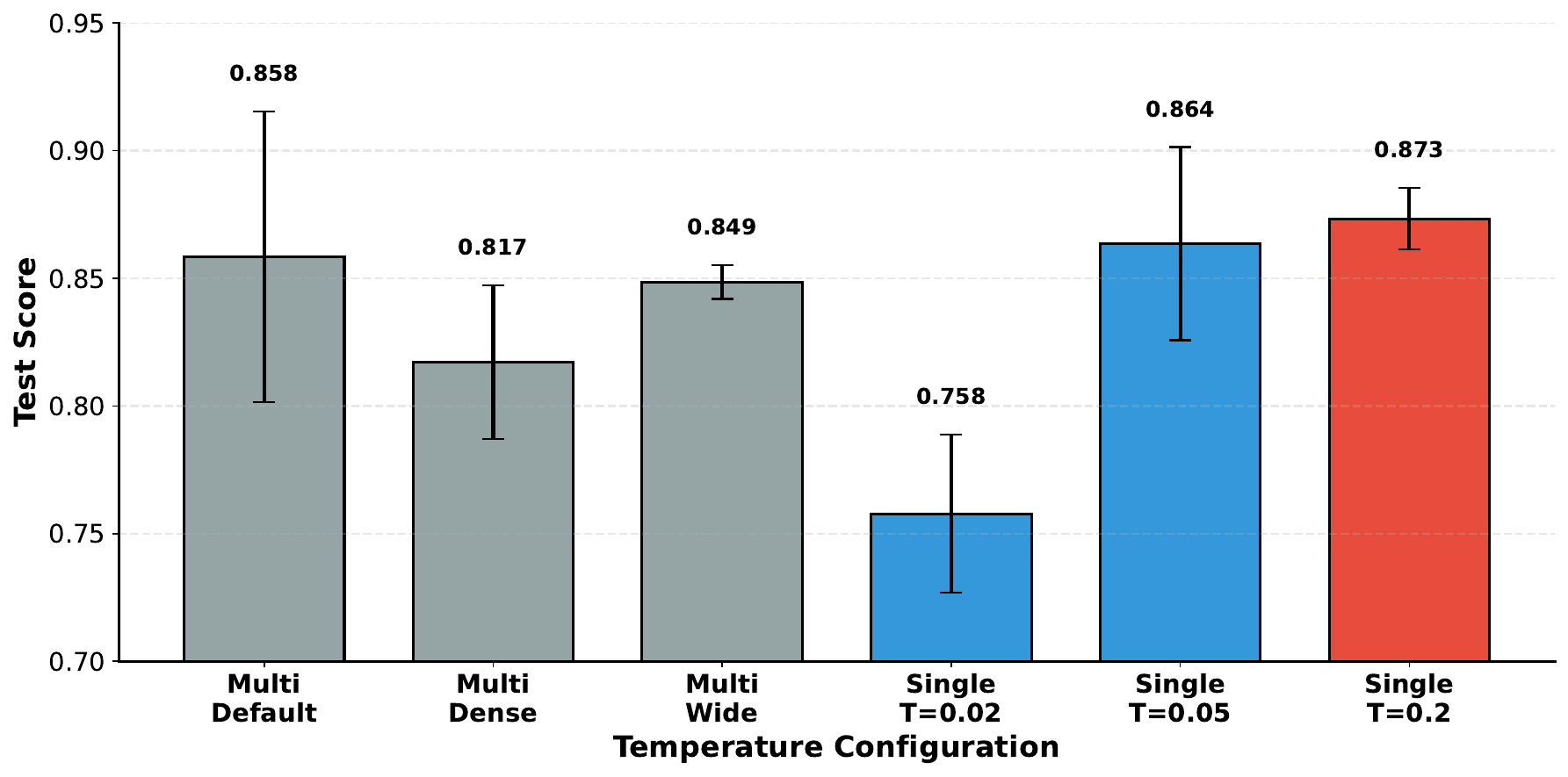}
\caption{Test performance across temperature configurations (mean $\pm$ std over 3 seeds, 50 rollouts each). Single temperature $T=0.2$ achieves the best performance with the lowest variance.}
\label{fig:ablation_temperature}
\end{figure}

Table~\ref{tab:ablation_temperature_results} and Figure~\ref{fig:ablation_temperature} present the results. Three key patterns emerge:

\paragraph{Single-temperature superiority.} Single-temperature configurations consistently outperform multi-temperature setups. The best single-temperature ($T=0.2$, score $0.873$) exceeds the best multi-temperature (\texttt{multi\_default}, score $0.858$) by $1.7\%$, indicating that a single well-chosen scale suffices to capture task-relevant geometric structure on PushT.

\paragraph{Temperature scale sensitivity.} Performance varies significantly across single-$T$ values: from $T=0.02$ to $T=0.05$ yields a $14.0\%$ gain ($0.758\to0.864$), while $T=0.2$ adds another $1.0\%$ ($0.864\to0.873$). Moderate-to-large temperatures better capture the global action-space geometry of this manipulation task.

\paragraph{Stability and diminishing returns.} \texttt{single\_T0p20} achieves both the highest mean performance and the lowest cross-seed variance (std $0.012$). In contrast, \texttt{multi\_default} exhibits significantly higher variance (std $0.057$) despite using the same temperature values, suggesting multi-scale aggregation introduces additional optimization complexity. Increasing temperature count further---\texttt{multi\_dense} uses 6 temperatures---degrades performance by $4.8\%$ relative to \texttt{multi\_default} ($0.817$ vs.\ $0.858$).

\paragraph{Recommendation.} Use single temperature $T=0.2$ as the default for manipulation tasks of similar complexity. For tasks with substantially different action-space geometry, first sweep single-temperature values in $[0.05,0.5]$ before considering multi-temperature configurations.

% -------------------------------------------------------------------
\subsection{Analysis of Loss Computation Modes}
% -------------------------------------------------------------------

The drift-field regression objective supports two modes: \emph{chunk mode} computes loss over the entire flattened action chunk ($S=Hd_a$), enforcing joint temporal coherence; \emph{step-wise mode} computes loss independently at each timestep ($S=d_a$) and averages over the horizon, reducing memory at the cost of inter-step dependencies. We compare the two modes with $G=8$, batch size $32$, temperature $\mathcal{R}=\{0.02,0.05,0.2\}$, and 200 training epochs (extended to ensure convergence; 6 total runs across $2\times3$ seeds).

\begin{table}[h]
\centering
\caption{Loss computation mode comparison. Chunk mode achieves $5.3\%$ higher performance with lower cross-seed variance.}
\label{tab:ablation_chunk_results}
\small
\begin{tabular}{lcc}
\toprule
\textbf{Mode} & \textbf{Mean Score} & \textbf{Std Score} \\
\midrule
\textbf{Chunk mode} ($S=Hd_a$, \texttt{per\_timestep\_loss=False}) & \textbf{0.890} & \textbf{0.015} \\
Step-wise mode ($S=d_a$, \texttt{per\_timestep\_loss=True}) & 0.845 & 0.021 \\
\bottomrule
\end{tabular}
\end{table}

Table~\ref{tab:ablation_chunk_results} and Figure~\ref{fig:ablation_chunk_vs_step} present the results.

\paragraph{Performance and stability.} Chunk mode achieves $5.3\%$ higher mean performance ($0.890$ vs.\ $0.845$) and lower cross-seed variance (std $0.015$ vs.\ $0.021$). The richer geometric structure available in the larger flattened space ($S=Hd_a$) provides a more stable and informative learning signal.

\paragraph{Temporal coherence.} The performance gap reflects temporal coherence preservation. Chunk mode enforces consistency across the entire predicted horizon through joint optimization in flattened space, while step-wise mode treats each timestep independently. For manipulation tasks requiring smooth action sequences, this joint optimization is critical.

\paragraph{Recommendation.} Use chunk mode (\texttt{per\_timestep\_loss=False}) as the default. Step-wise mode may be considered when memory constraints prohibit full-chunk tensor operations, but the $5.3\%$ performance cost should be carefully factored in.

% -------------------------------------------------------------------
\subsection{Training Efficiency Comparison}
% -------------------------------------------------------------------

Beyond hyperparameter sensitivity, the one-step generation paradigm yields a fundamental efficiency advantage over iterative diffusion methods. Table~\ref{tab:training_efficiency_comparison} compares training and inference requirements on PushT.

\begin{table}[h]
\centering
\caption{Training and inference efficiency comparison on the PushT low-dimensional task.}
\label{tab:training_efficiency_comparison}
\small
\begin{tabular}{lcc}
\toprule
\textbf{Method} & \textbf{Training Epochs} & \textbf{Inference NFE} \\
\midrule
Diffusion Policy & 4,500 & 100 \\
\textbf{Drift-Based Policy (Ours)} & \textbf{100} & \textbf{1} \\
\bottomrule
\end{tabular}
\end{table}

The $45\times$ reduction in training epochs stems from eliminating the iterative score-matching objective: diffusion models require multi-step denoising during both training and inference, whereas DBP learns a direct latent-to-action mapping in a single forward pass. This translates to proportional savings in wall-clock time and computational resources. Despite training for $45\times$ fewer epochs, our configurations achieve competitive task performance, demonstrating that drift-field regression provides a more efficient learning signal than iterative score matching.

% -------------------------------------------------------------------
\subsection{Complete Hyperparameter Reference}
% -------------------------------------------------------------------

For full reproducibility, Tables~\ref{tab:hyperparameters_stage1} and~\ref{tab:hyperparameters_architecture} list all hyperparameters used in the DBP experiments. Tuned hyperparameters were selected via grid search on PushT (3 seeds per configuration): $G=4$ achieved the highest mean performance ($0.887$) with acceptable cost; single temperature $T=0.2$ offered both higher mean performance ($0.873$) and lower variance (std $0.012$) than all multi-scale alternatives; chunk mode was chosen despite higher memory usage due to its $5.3\%$ absolute performance advantage.

\begin{table}[h]
\centering
\caption{DBP hyperparameters. All values are held constant across experiments unless explicitly varied in the corresponding analysis.}
\label{tab:hyperparameters_stage1}
\small
\begin{tabular}{lccc}
\toprule
\textbf{Hyperparameter} & \textbf{Value} & \textbf{Tuned?} & \textbf{Tuning Range} \\
\midrule
\multicolumn{4}{l}{\textit{Drift-Field Method}} \\
Hypothesis count $G$ & 4 & Yes & $\{1, 2, 4, 8, 16\}$ \\
Temperature set $\mathcal{R}$ & $\{0.2\}$ & Yes & Single: $\{0.02, 0.05, 0.2\}$ \\
 & & & Multi: various combinations \\
Negative reference count $C_n$ & 0 & No & -- \\
Positive reference count $C_p$ & 1 & No & -- \\
Scale normalization floor $\epsilon_s$ & $10^{-6}$ & No & -- \\
Force normalization floor $\epsilon_f$ & $10^{-6}$ & No & -- \\
Loss computation mode & chunk & Yes & $\{\text{chunk}, \text{step-wise}\}$ \\
\midrule
\multicolumn{4}{l}{\textit{Training Configuration}} \\
Batch size $B$ & 32 & Yes & $\{16, 32, 48, 64, 96\}$ \\
Learning rate & $10^{-4}$ & Yes & $\{10^{-5}, 10^{-4}, 10^{-3}\}$ \\
Optimizer & AdamW & No & -- \\
Adam $\beta_1$ & 0.95 & No & -- \\
Adam $\beta_2$ & 0.999 & No & -- \\
Weight decay & $10^{-6}$ & No & -- \\
Gradient clipping & 1.0 & No & -- \\
Training epochs & 100 & No & -- \\
LR warmup steps & 500 & No & -- \\
LR schedule & Constant after warmup & No & -- \\
EMA decay & 0.9999 & No & -- \\
EMA power & 0.75 & No & -- \\
\midrule
\multicolumn{4}{l}{\textit{Action Prediction}} \\
Prediction horizon $H$ & 16 & No & -- \\
Execution steps $H_e$ & 8 & No & -- \\
Observation history $T_o$ & 2 & No & -- \\
\bottomrule
\end{tabular}
\end{table}

\begin{table}[h]
\centering
\caption{Network architecture hyperparameters. The generator uses a conditional U-Net architecture with 1D convolutions.}
\label{tab:hyperparameters_architecture}
\small
\begin{tabular}{ll}
\toprule
\textbf{Component} & \textbf{Configuration} \\
\midrule
\multicolumn{2}{l}{\textit{Generator (Conditional U-Net 1D)}} \\
Latent dimension & 256 \\
Down-sampling channels & [512, 1024, 2048] \\
Up-sampling channels & [2048, 1024, 512] \\
Kernel size & 5 \\
Normalization & GroupNorm (8 groups) \\
Activation & SiLU \\
Dropout & 0.1 \\
Attention layers & At middle resolution \\
Time embedding dim & 256 \\
Condition embedding dim & 256 \\
\bottomrule
\end{tabular}
\end{table}

\section{Qualitative Visualization of Policy Execution}
\label{sec:policy_visualization}

\subsection{Visualization Scope and Purpose}

To provide qualitative evidence of the Drift-Based Policy's execution quality, we visualize rollout trajectories on representative tasks from the Adroit and Meta-World benchmarks. These visualizations complement the quantitative success rate metrics reported in the main paper by illustrating the temporal coherence and spatial precision of the learned policies under point-cloud observations.

The visualizations capture key aspects of policy behavior: (i) smooth action progression throughout task execution, (ii) precise manipulation of objects in 3D space, and (iii) successful task completion under the strict 1-NFE inference constraint. Each visualization sequence shows temporally sampled frames from a single successful rollout, demonstrating that the one-step generation paradigm maintains control quality without iterative refinement at deployment.

\begin{figure*}[p]
\centering
\includegraphics[width=0.98\textwidth]{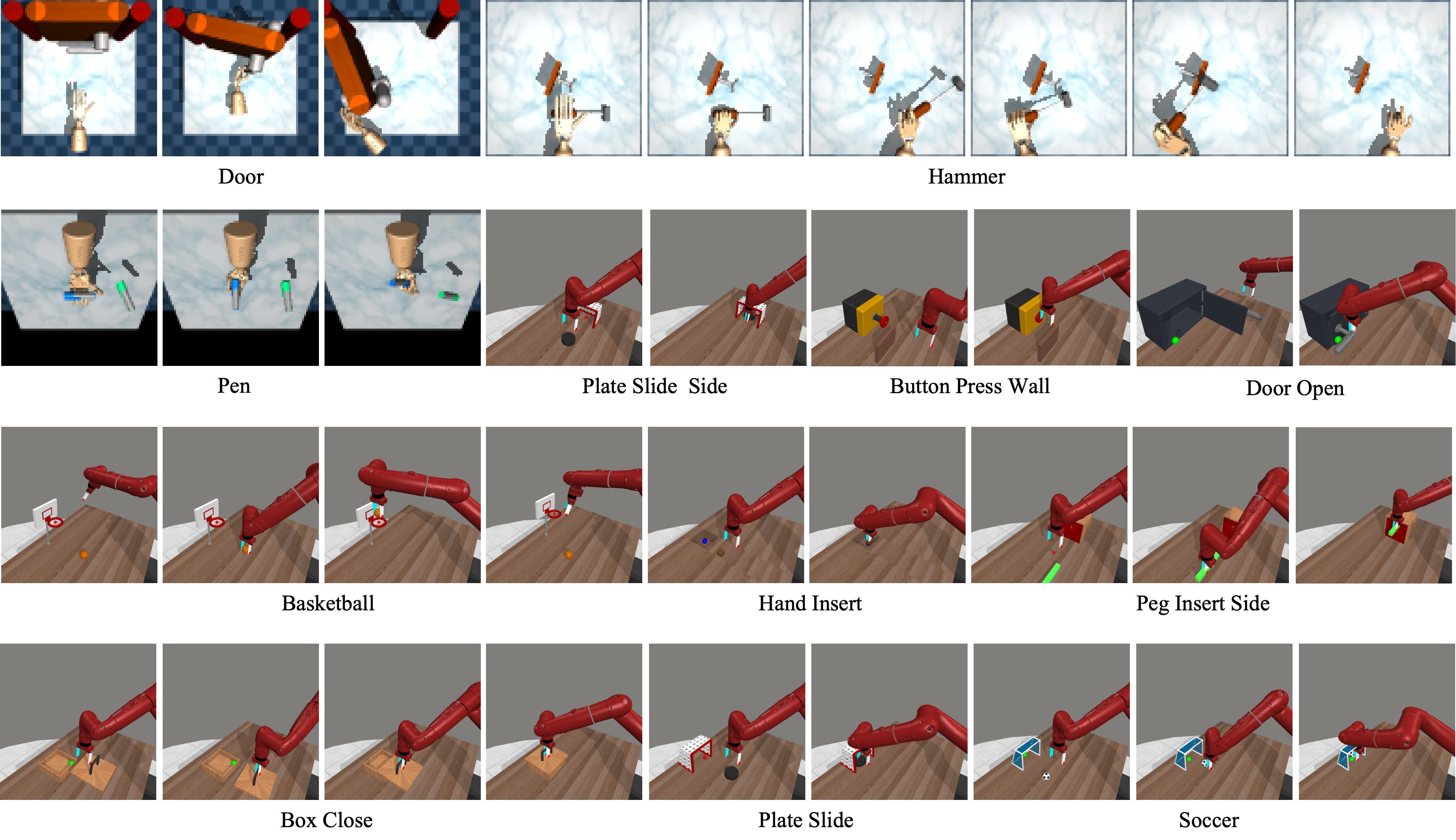}
\caption{Drift-Based Policy execution on Adroit and Meta-World tasks (Part 1). Frames show temporally sampled states from successful rollouts under 1-NFE inference, demonstrating smooth manipulation and precise object control.}
\label{fig:policy_visualization_part1}
\end{figure*}

\begin{figure*}[p]
\centering
\includegraphics[width=0.98\textwidth]{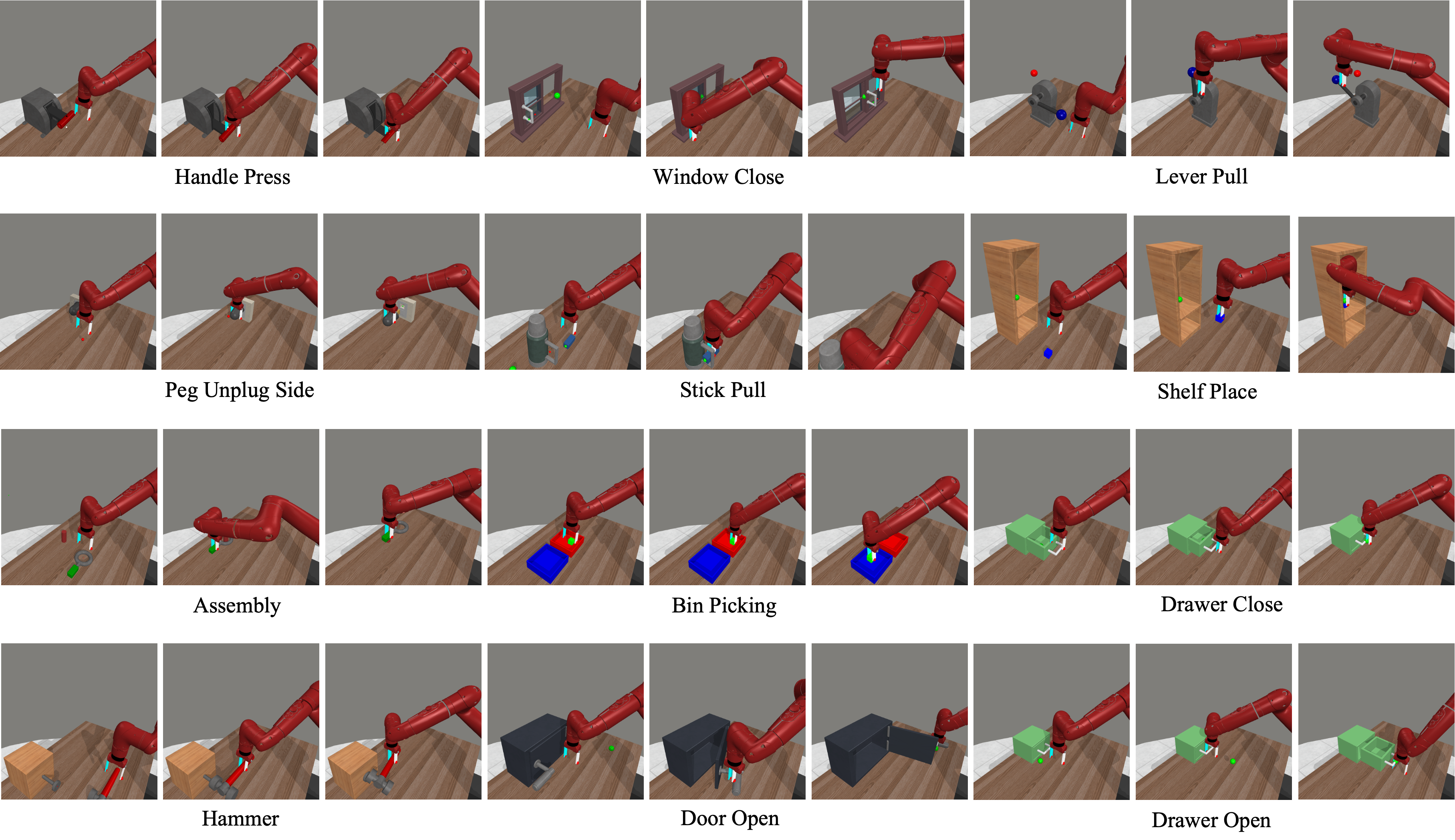}
\caption{Drift-Based Policy execution on Adroit and Meta-World tasks (Part 2). The temporal sequences demonstrate stable execution and coherent action progression without iterative correction at inference time.}
\label{fig:policy_visualization_part2}
\end{figure*}

\subsection{Interpretation of Visualization Sequences}

Each visualization panel presents a temporal sequence of frames sampled uniformly from a complete task execution. The sequences are selected to represent diverse manipulation scenarios across different task difficulties and object configurations.

The visualizations reveal several consistent patterns across tasks. First, the policy maintains smooth spatial trajectories without abrupt discontinuities, indicating that the internalized drift-field refinement successfully produces coherent action sequences. Second, object manipulation exhibits precise spatial control, with the end-effector consistently achieving target configurations despite the challenging point-cloud observation modality. Third, the temporal progression from initial state to goal state demonstrates stable execution without the need for iterative correction at inference time.

These qualitative observations align with the quantitative results in the main paper, where DBP achieves an 88.4\% success rate across 37 point-cloud manipulation tasks. The visualizations provide complementary evidence that high success rates are accompanied by smooth, precise execution trajectories, supporting the claim that native one-step generation can maintain control quality while eliminating multi-step inference overhead.